\documentclass[preprint,10pt,authoryear]{./layout/elsarticle}

\usepackage{amssymb}
\usepackage{amsmath}
\usepackage{accents}
\usepackage{mathtools}
\usepackage{booktabs}
\usepackage[flushleft]{threeparttable}

\usepackage{graphicx}
\usepackage{hyperref}
\usepackage{caption}
\usepackage{subcaption}
\usepackage{siunitx}
\usepackage{float}
\usepackage{array}
\usepackage{tikz}

\usetikzlibrary{shapes.geometric, arrows, calc}
\tikzstyle{header} = [coordinate]
\tikzstyle{element} = [rectangle, rounded corners, 
minimum width=2.5cm, 
minimum height=0.7cm,
text centered, 
draw=black]
\tikzstyle{arrow} = [thick,->,>=stealth]

\DeclareCaptionFormat{myformat}{#1#2#3\hrulefill}
\journal{Biomedical Signal Processing and Control}

\begin{document}

\begin{frontmatter}

\title{A CNN-Transformer for Classification of Longitudinal 3D MRI Images -- A Case Study on Hepatocellular Carcinoma Prediction}







\author[1,3]{Jakob Nolte}
\author[1]{Maureen M. J. Guichelaar}
\author[2]{Donald E. Bouman}
\author[3]{Stéphanie M. van den Berg}
\author[3]{Maryam Amir Haeri}

\affiliation[1]{organization={Department of Gastroenterology and Hepatology, Medisch Spectrum Twente}, 
                city={Enschede},
                country={The Netherlands}}

\affiliation[2]{organization={Department of Radiology, Medisch Spectrum Twente}, 
                city={Enschede},
                country={The Netherlands}}

\affiliation[3]{organization={Department of Learning, Data Analytics and Technology, University of Twente}, 
                city={Enschede},
                country={The Netherlands}}

\begin{abstract}
Longitudinal MRI analysis is crucial for predicting disease outcomes, particularly in chronic conditions like hepatocellular carcinoma (HCC), where early detection can significantly influence treatment strategies and patient prognosis. Yet, due to challenges like limited data availability, subtle parenchymal changes, and the irregular timing of medical screenings, current approaches have so far focused on cross-sectional imaging data. To address this, we propose HCCNet, a novel model architecture that integrates a 3D adaptation of the ConvNeXt CNN architecture with a Transformer encoder, capturing both the intricate spatial features of 3D MRIs and the complex temporal dependencies across different time points.

HCCNet utilizes a two-stage pre-training process tailored for longitudinal MRI data. The CNN backbone is pre-trained using a self-supervised learning framework adapted for 3D MRIs, while the Transformer encoder is pre-trained with a sequence-order-prediction task to enhance its understanding of disease progression over time. We demonstrate the effectiveness of HCCNet by applying it to a cohort of liver cirrhosis patients undergoing regular MRI screenings for HCC surveillance. Our results show that HCCNet significantly improves predictive accuracy and reliability over baseline models, providing a robust tool for personalized HCC surveillance.

The methodological approach presented in this paper is versatile and can be adapted to various longitudinal MRI screening applications. Its ability to handle varying patient record lengths and irregular screening intervals establishes it as an invaluable framework for monitoring chronic diseases, where timely and accurate disease prognosis is critical for effective treatment planning.
\end{abstract}


\end{frontmatter}

\section{Introduction}
\label{sec:introduction}
Predicting disease outcomes based on longitudinal MRI scans is crucial for improving patient management. This is especially important in conditions like hepatocellular carcinoma (HCC), where early detection of tissue changes and subsequent monitoring can significantly impact treatment decisions and prognosis \citep{parikh2020biomarkers}. Longitudinal MRI analysis enables the capture of disease progression over time, providing a comprehensive view of how pathological changes evolve \citep{gong2024individualised}. However, despite its potential, the development of effective predictive models that can fully leverage the temporal information in longitudinal MRI data remains a significant challenge \citep{jin2021predicting}.

One of the major challenges lies in accurately detecting early signs of disease progression from MRI scans, particularly in identifying subtle changes in the underlying parenchyma that may eventually lead to a malignancy. These challenges are further compounded by the need to accurately capture complex temporal dependencies across multiple time points while preserving the spatial integrity of the 3D MRI data \citep{tang2021self}. Traditional approaches often simplify this task by extracting 2D slices from 3D MRI scans and applying convolutional neural networks (CNNs) pre-trained on large-scale natural image datasets \citep{kumar2024improving}. However, this method has limitations, particularly in the medical domain where datasets are often small, and the differences between natural and medical images—such as texture, contrast, and noise—significantly impact the transferability of pre-trained models \citep{zeng2021positional}.

The scarcity of large, labeled datasets in medical imaging exacerbates these challenges, making it difficult to train models that generalize well across different patient populations. As a result, pre-training approaches have become crucial in the medical domain, helping to improve model performance by enabling the network to learn robust features from limited data \citep{tang2021self}. However, many existing approaches have not fully explored the potential of these techniques in the context of longitudinal MRI analysis, particularly for diseases that require monitoring over extended periods \citep{parikh2020biomarkers}. Likewise, most deep learning applications are aimed at detecting a malignancy, resulting in a lack of studies focusing on detecting tissue changes before the tumor actually develops. 

In response to these limitations, we propose HCCNet, a novel model architecture designed specifically for longitudinal 3D MRI analysis. HCCNet integrates a 3D adaptation of the ConvNeXt architecture with a Transformer encoder, allowing it to capture both the spatial features of 3D MRIs and the temporal dependencies between different scans \citep{gong2024individualised}. Unlike previous approaches that rely on 2D slices or disregard the irregular intervals between scans, HCCNet is designed to handle the complexities of longitudinal data, making it better suited for tasks like predicting disease progression \citep{jin2021predicting}.

To address the challenges of early detection and small datasets, HCCNet incorporates a robust pre-training strategy. The CNN backbone is pre-trained using an extended version of the DINO framework, adapted to 3D MRI, which treats different MRI sequences as natural augmentations of the same image \citep{tang2021self}. This self-supervised learning approach allows the model to learn from the inherent variability in MRI data, even when labeled examples are scarce \citep{kumar2024improving}. Additionally, the Transformer encoder is pre-trained with a sequence-order-prediction task inspired by techniques used in natural language processing, which helps the model to understand the progression of disease over time and to make accurate predictions despite the irregular timing of medical screenings \citep{zeng2021positional}.

Our study specifically applies HCCNet to the task of predicting HCC development in liver cirrhosis patients, based on longitudinal MRI scans. By fine-tuning the pre-trained model on this dataset, we demonstrate the model's ability to predict whether a patient will develop HCC at the next examination \citep{parikh2020biomarkers}. This application not only serves as a showcase of HCCNet’s capabilities but also addresses a critical need in personalized HCC surveillance of cirrhotic patients, where timely and accurate predictions can significantly influence clinical outcomes. Furthermore, the methodological approach presented in this paper can also be considered for predicting other conditions that, similar to HCC, benefit from monitoring changes over time. Thus, HCCNet provides a versatile framework that can be adapted to various longitudinal medical imaging tasks, potentially improving predictive accuracy and patient outcomes across a range of diseases. All code for the proposed method is available at: \href{https://github.com/jmnolte/HCCNet}{github.com/jmnolte/HCCNet}.

\section{Related Work}
\label{sec:related_work}

\subsection{Modeling Approaches in Longitudinal Medical Image Analysis}
\label{sec:early_diagnosis}
To date, deep learning applications to medical imaging still predominately consider cross-sectional instead of repeated imaging data. However,  physicians usually assess the progression of disease or the effect of treatment by comparing patients' current to previous imaging records. As a result, longitudinal imaging has long been established as the gold standard in many branches of oncology, enticing recent studies to extend the analysis of medical imaging to the analysis of imaging data across time. 

Three main methodological frameworks have emerged from these studies. Among them, fully convolutional neural networks (CNNs) model the within person change across time by stacking the image screenings from multiple time points into a multidimensional image and feeding the single joint image through the network \citep{gaoPredictionSoftTissue2021a, dammuDeepLearningPrediction2023}. While the approach is relatively simple to implement, it has been shown to not fully explore the relationship between time points \citep{dammuDeepLearningPrediction2023}. Studies investigating treatment effects have therefore increasingly employed Siamese networks \citep{jinPredictingTreatmentResponse2021, dammuDeepLearningPrediction2023}. These networks constitute a parallel architecture, where both parts receive different inputs but share the same set of parameter weights. Applied to treatment response analysis, pre- and post-treatment images are thus fed independently through the parallel network architecture and the final embedding is obtained by concatenating the individual vector representations. Despite promising results, Siamese networks are only applicable to the comparison of pairs of time points, making them inadequate for the analysis of long-term disease trajectories. 

To mitigate this shortcoming, our study utilizes a hybrid model architecture, which is the third most common framework in longitudinal medical image analysis. Contrary to the other approaches, these models not only rely on a CNN backbone to embed the raw image screenings in high-dimensional feature space but also employ a sequential model to process the sequence of embedded image representations. Hybrid models can be trained end-to-end, consecutively, or using a pre-trained CNN feature extractor and typically employ a recurrent neural network (RNN), long-short-term memory (LSTM) or Transformer model to process the sequence of image embeddings. Compared to the aforementioned approaches, hybrid models can handle patient records of varying time points, making them especially suited for the continued surveillance of patients at high-risk of disease or after the onset of treatment. As such, they have been successfully employed in response prediction to radiotherapy \citep{leeDeepLearningDriven2022, wangPredictingEvolutionLung2019, wangPredictingSpatialEsophageal2020, xuDeepLearningPredicts2019} and endothelial therapy \citep{luDeepLearningPrediction2021} as well as in breast cancer risk prediction \citep{dadsetanDeepLearningLongitudinal2022}. Analyzing ultrasound (US) images from 619 cirrhotic subjects, \cite{zhangSpatiotemporalAttentionEarly2022} even applied them to HCC risk prediction, obtaining a moderate area under the receiver operating characteristic curve (AUC) of 0.775. 

However, all studies employing hybrid models rely, to the best of our knowledge, on 2D imaging modalities like ultrasound or the extraction of 2D slices from 3D MRI or CT volumes. Likewise, many of them disregard the irregular time intervals between image screenings. In this study, we therefore propose a spatio-temporal 3D CNN-Transformer hybrid model that can capture the long-range dependencies among longitudinal MRI scans and utilizes time interval based positional encodings to account for the irregular examination intervals between image screenings.

\subsection{Pre-Training Approaches in Longitudinal Medical Image Analysis}
\label{sec:pretraining_related}
While the temporal relationship across medical image screening is increasingly investigated, pre-training approaches for longitudinal medical image analysis remain scarcely explored. As a result, many deep learning studies analyzing medical images over time continue to extract 2D slices from 3D medical images to leverage CNN model weights pre-trained on large scale natural image datasets. However, the usefulness of transfer learning from natural images remains fiercely disputed \citep{zhangAdvancing3DMedical2022, huangSelfsupervisedLearningMedical2023, el-noubyAreLargescaleDatasets2021} and our study also employs a 3D rather than a 2D CNN. Utilizing the potential abundance of unlabeled data, some longitudinal medical imaging studies have therefore started to apply self-supervised pre-training approaches to enhance downstream model performance. 

Among them, contrastive learning approaches remain the most commonly applied pre-training framework. These approaches build on the assumption that alterations caused by image transformations do not alter the image's semantic meaning. As such, they randomly derive different views from the same input image \citep{chenSimpleFrameworkContrastive2020} or from the same patient at different time points and learn rich contextual feature embeddings by matching the view's embedded representation to the one obtained with the other part of a Siamese network or a momentum encoder \citep{grillBootstrapYourOwn2020}. \cite{jamaludinSelfsupervisedLearningSpinal2017}, for example, pre-trained a Siamese CNN on baseline and follow-up image pairs by minimizing the distance between view pairs from the same patient and maximizing the distance between images from different patients. \cite{rivailModelingDiseaseProgression2019} extended this approach such that it accounts for irregularly sampled data. Instead of simply comparing view pairs from different patients, they additionally considered the time interval between image screenings from the same patient for their loss calculation. 

While above approaches proved successful in their respective downstream application, they do not provide a framework for patient records of varying lengths or hybrid model approaches in general. We therefore propose a novel pre-training framework for longitudinal 3D MRI scans that captures both spatial and temporal dependencies both within and between patients. The framework borrows from the contrastive DINO framework proposed by \cite{caronEmergingPropertiesSelfSupervised2021} for pre-training of the CNN backbone and adopts a simple sequence-order-prediction task inspired by \cite{lanALBERTLiteBERT2020} and \cite{renRAPTPretrainingTimeAware2021} for pre-training of the Transformer encoder. Further details on the study's pre-training implementation are discussed in section \ref{sec:pretraining}.

\section{Methodology}
\label{sec:methodology}

\subsection{Problem Formulation}
\label{sec:problem_formulation}
This study aims to define a mapping function $f$ that for each liver cirrhosis patient learns to predict whether the patient will be diagnosed with HCC at the next examination given a set of MR images from past patient visits. Since the frequency of liver screenings is typically determined by the patient's unique condition, some patients may undergo more frequent screenings than others with more or less regular time intervals between examinations. As such, the data of each patient $p \in \{1, 2, ..., P\}$ corresponds to a vector of unique length and is denoted by $D^p = \{i^{(p,1)}, i^{(p,2)}, ..., i^{(p,t)}; t_{HCC}^p\}$, where ${i^{(p,t)}}$ is the patient's MRI examination at time point $t$ and $t_{HCC}^p$ is the patient's date of diagnosis which is set to None if HCC was not diagnosed within $t$. For each patient with $t_{HCC}^p \neq \text{None}$, patient records are subset during fine-tuning such that the patient's last considered observation $i^{(p,t)}$ has time point $t < t_{HCC}^p$. Likewise, the label for the sequence of patient visits in subset $V^{(p,t)} = \{i^{(p,1)}, i^{(p,2)}, ..., i^{(p,t)}\} \subseteq D^{(p,t)}$ is set to:
\begin{equation}
    \textbf{y}_p=\begin{cases}
        0, & \text{if}\ t_{HCC}^p = \text{None},\\
        1, & \text{otherwise}.
  \end{cases}
\end{equation}

\subsection{Overall Architecture}
\label{sec:architecture}
\begin{figure}[t!]
    \centering
    \caption{Proposed Modeling Approach}
    \includegraphics[width=\textwidth]{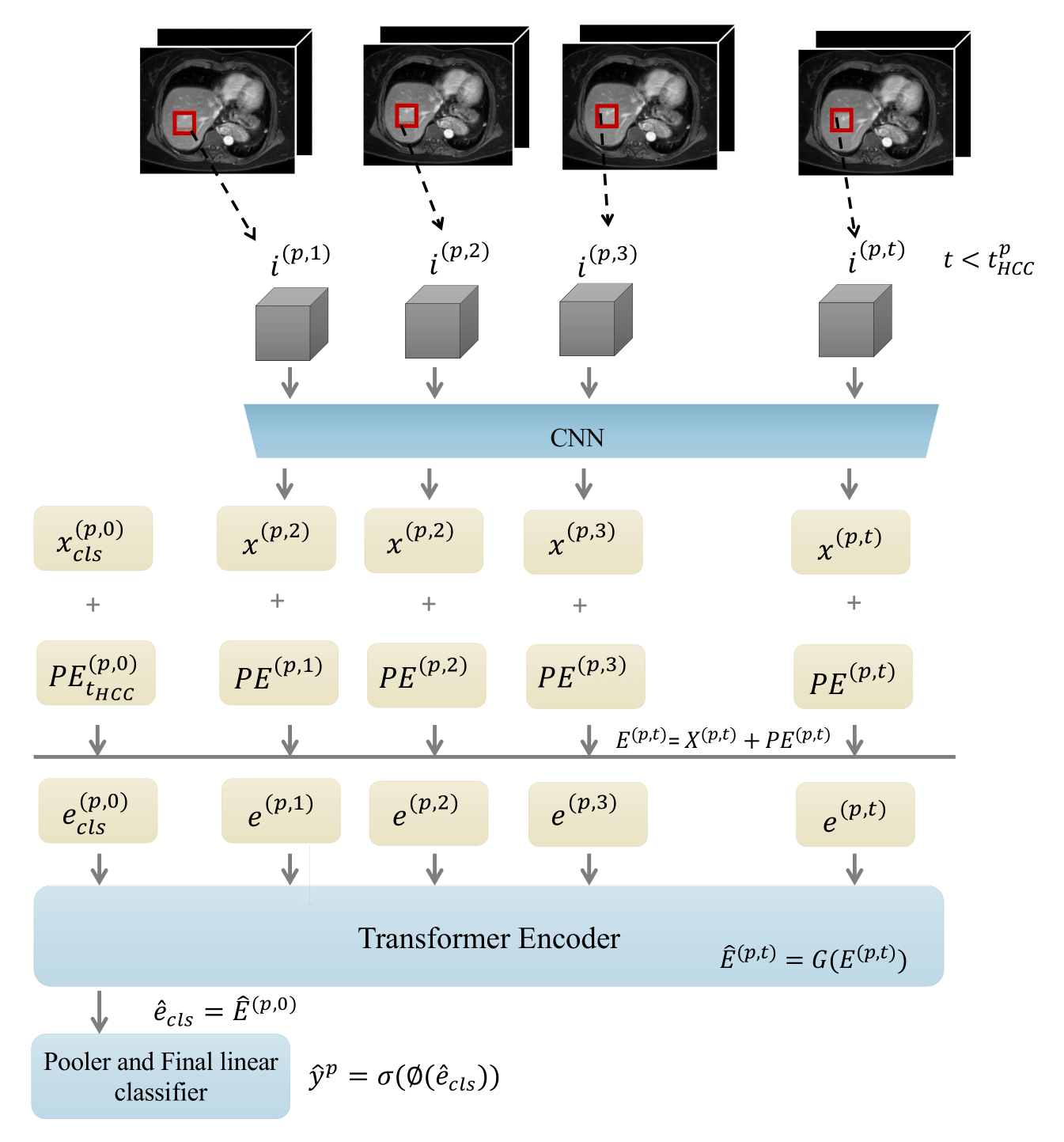}
    \label{fig:arch}
    \hrule
    \begin{minipage}{0.99\textwidth}
        \vspace{1mm}
        \footnotesize
        \emph{Note}: The figure presents the modeling approach for an exemplary patient with longitudinal MRI screening. Raw MRI's are fed through the CNN backbone before adding positional encodings to the image embeddings. The resulting embeddings are fed through the Transformer encoder and the last hidden state of the [cls] token used for prediction.
    \end{minipage}
\end{figure}

To handle the longitudinal nature of the patient records, HCCNet is a CNN-Transformer model based on a 3-dimensional adaptation of ConvNeXt \citep{liuConvNet2020s2022} and the original Transformer encoder architecture \citep{vaswaniAttentionAllYou2023}. We utilize the CNN backbone to embed the raw MR images in higher-dimensional feature space and the Transformer encoder to process the sequence of embedded image representations. A high level overview of the proposed modeling approach is provided in Figure \ref{fig:arch}. 

\textbf{CNN Feature Extractor.} For the CNN backbone, we adapt the ConvNeXt architecture proposed in \citep{liuConvNet2020s2022}. More specifically, we replace the 2-dimensional depthwise convolutional layers in ConvNeXt with their 3-dimensional counterpart and reduce the proposed kernel size from $k = 7$ to $k = 3$. Additionally, we reimplement the patchify layer in ConvNeXt with a smaller kernel size and stride for the convolutional layer resulting in a 3x rather than 4x downsampling of the input images and thus accommodating for the lower input size typically encountered in volumetric data. More details on the study's 3D implementation of ConvNeXt are presented in \ref{sec:appendix_conv_layer}.

The CNN feature extractor takes the MR images as inputs and returns a high-dimensional vector representation of the raw images. The vector representation can be denoted by:
\begin{equation}
    \textbf{x}^{(p,t)} = F(i^{(p,t)}),
\end{equation}

\noindent where $F$ is the CNN backbone, $\textbf{x}^{(p,t)} \in \mathbb{R}^H$ is the resulting embedding for patient $p$ at time point $t$, and $H = \text{dim}(\textbf{x}^{(p,t)})$.

\textbf{Transformer Encoder.} The Transformer encoder is based on the Pre-LN implementation \citep{xiongLayerNormalizationTransformer2020} of the original Transformer encoder architecture \citep{vaswaniAttentionAllYou2023} and is implemented using $L$ encoder layers with a hidden size of $H = \text{dim}(\textbf{x}_{(p,t)})$ and $A = \frac{H}{128}$ self-attention heads. To model the disease trajectory over time, we concatenate the image embeddings obtained from the CNN feature extractor and prepend a learnable classification token [cls] yielding the following sequence of image representations:
\begin{equation}
    \textbf{X}^{(p,t)} = [\textbf{x}_\text{cls}^{(p,0)}, \textbf{x}^{(p,1)}, \textbf{x}^{(p,2)}, ..., \textbf{x}^{(p,t)}] \in \mathbb{R}^{(t+1) \times H},
\end{equation}

\noindent where $\textbf{x}^{(p,t)}$ is the previously derived image embedding and $\textbf{x}_\text{cls}^{(p,0)}$ is the [cls] token of patient $p$.

Following that, we embed the input's positional information using fixed sinusodial positional encodings. To account for the irregularly spaced time intervals between patients' visits, we substitute the conventionally employed position in the sequence with the distance (i.e., the time in months) between the date of diagnosis and all previous examinations for patients with developing HCC or the patient's last registered MRI screening and all previous examinations for patients without developing HCC respectively. In addition, we take the square root of the time rather than the non-transformed distance since we assume a non-linear decay in importance of previous observations for future prediction. In other words, we assume that the distance between observations registered, e.g., 5 and 6 years prior to diagnosis is less salient for prediction than the distance between observations taken, for example, 3 months and 1 year before diagnosis. Given a vector of time points $\textbf{t}^p_i = (t_1, t_2, ..., t_i; t^p_{\text{HCC}})$ for patient $p$, where $t_{HCC}^p$ is the date of diagnosis, we thus compute the vector of time intervals: 
\begin{equation}
    \boldsymbol{\Delta}\textbf{t}^p_i=\begin{cases}
        \sqrt{t_{i+1} - t_i}, & \text{if}\ t_{\text{HCC}}^p = \text{None},\\
        \sqrt{t_{\text{HCC}}^p - t_i}, & \text{otherwise},
  \end{cases}
\end{equation}

\noindent prepend $\boldsymbol{\Delta}\textbf{t}^p_{\text{cls}} = 0$ for the classification [cls] token, and, finally, obtain positional encodings using sine and cosine functions of different frequencies:
\begin{align}
    \begin{split}
        \textbf{PE}_{(\boldsymbol{\Delta}\textbf{t}^p_i, 2k)} &= \text{sin}(\sqrt{\boldsymbol{\Delta}\textbf{t}^p_i}/10000^{2k/H}), \\
        \textbf{PE}_{(\boldsymbol{\Delta}\textbf{t}^p_i, 2k+1)} &= \text{cos}(\sqrt{\boldsymbol{\Delta}\textbf{t}^p_i}/10000^{2k/H}),
    \end{split}
\end{align}

\noindent where $k$ is the dimension and $H$ is the model's hidden size. After encoding the temporal information, we add the positional encodings to the embedding matrix obtaining the final input to the Transformer encoder:
\begin{equation}
    \textbf{E}^{(p,t)} = \textbf{X}^{(p,t)} + \textbf{PE}^{(p,t)}  \in \mathbb{R}^{(t+1) \times H}.
\end{equation}

We feed the input sequence through the encoder $G$ and select the last hidden state of the [cls] token. The operation can be formulated as:
\begin{equation}
    \hat{\textbf{E}}^{(p,t)} = G(\textbf{E}^{(p,t)}), 
\end{equation}

\noindent where $\hat{\textbf{E}}^{(p,t)} \in \mathbb{R}^{(t+1) \times H}$ is the model's final hidden state and $\hat{\textbf{e}}_\text{cls} = \hat{\textbf{E}}^{(p,0)} \in \mathbb{R}^H$ is the last hidden state of the classification token. To enhance the representation embedded in the [cls] token's final hidden state, we feed it through a pooling layer $\phi$ consisting of a linear layer followed by a Tanh activation, a normalization layer, and the linear classification layer, before finally applying the sigmoid function to obtain the predicted probability of HCC development. The final model output is thus denoted by:
\begin{equation}
    \hat{\textbf{y}}^p = \sigma(\phi(\hat{\textbf{e}}_\text{cls})),
\end{equation}

\noindent where $\hat{\textbf{y}}^p$ is the model's predicted probability, $\phi$ is the model's pooling layer, and $\sigma$ is the sigmoid activation function.

\subsection{Architectural Variants}
\label{sec:variants}
In addition to our base model, called HCCNet-Pico, we introduce three variants of HCCNet, which are versions of about 0.5x, 2x, and 4x the base model's size respectively. The four model variations are: 
\begin{itemize}
    \item HCCNet-F: $C = 48$, $B$ = \{2, 2, 6, 2\}, $H = 384$, $A = 3$, $L = 4$
    \item HCCNet-P: $C = 64$, $B$ = \{2, 2, 6, 2\}, $H = 512$, $A = 4$, $L = 4$
    \item HCCNet-N: $C = 80$, $B$ = \{2, 2, 8, 2\}, $H = 640$, $A = 5$, $L = 6$
    \item HCCNet-T: $C = 96$, $B$ = \{3, 3, 9, 3\}, $H = 768$, $A = 6$, $L = 6$
\end{itemize}

\noindent where $C$ represents the number of channels in the ConvNeXt backbone, $B$ are the number of ConvNeXt blocks, $H$ is the Transformer encoder's hidden dimension, $L$ are the number of encoder layers, and $A$ are the number of self-attention heads. The naming convention follows ConvNeXt's original implementation in \citep{liuConvNet2020s2022}. As such, F, P, N, and T denote femto, pico, nano, and tiny with 12.4, 22.0, 45.9, and 72.4 million parameters respectively.

\subsection{Pre-Training}
\label{sec:pretraining}
When training neural networks, task agnostic pre-training has been repeatedly shown to improve performance on downstream tasks. Additionally, subset $V^{(p,t)}$ contains only roughly 75\% of observations included in superset $D^{(p,t)}$. To fully leverage all available data and improve the representations learned during training, we therefore design two (self-)supervised pre-training tasks for both the CNN backbone and the Transformer encoder prior to training on our downstream task.

\textbf{CNN Feature Extractor.} For the CNN backbone, this study extends the self-supervised DINO pre-training framework described in \citep{caronEmergingPropertiesSelfSupervised2021} to 3D MR images. Accordingly, we define a student $F_s$ and a teacher network $F_t$, where the teacher is built as an exponential moving average of the student and we learn to match the output probability distributions of the networks (i.e., $p_{{F}_s}(i^{(p,t)})$ and $p_{{F}_t}(i^{(p,t)})$ respectively) by minimizing the cross-entropy loss w.r.t. the parameters of the student network ${\theta_F}_s$:
\begin{equation}
    \min_{{\theta_F}_s} - \sum_{i \in \{i_g^{(p,t)}\}} \sum_{i^\prime \in \{i_{g}^{(p,t)}, i_l^{(p,t)}\}} p_{{F}_t}(i) \text{log}(p_{{F}_s}(i^\prime)),
\end{equation}

\noindent where $i_g^{(p,t)}$ and $i_l^{(p,t)}$ are the global and local views, which are cropped from the original image and used as input to the CNN backbone. Note that in our implementation, we use a 3D crop of shape $s = 72^3$ for the global and $s = 48^3$ for the local views. Besides that, our implementation also proposes a novel augmentation strategy specific to MR images. Instead of randomly adjusting the brightness, contrast, saturation and hue of an image as in the original DINO paper \citep{caronEmergingPropertiesSelfSupervised2021}, we leverage the abundance of natural variation in MRI data. Specifically, we treat the different sequences of an MRI as natural augmentations of the same image. For an image view cropped from a Diffusion-Weighted MRI (DW-MRI) with diffusion coefficient $b \in \{0, 150, 400, 800\}$, we thus randomly sample one series out of the sequence of available MRI scans and teach the model to identify the same patient under various conditions. More details on the study's data augmentation strategy are provided in \ref{sec:appendix_augmentation}.

\textbf{Transformer Encoder.} To improve the encoder's understanding of disease progression, this study employs a simple binary sequence-order-prediction (SOP) task inspired by \citep{lanALBERTLiteBERT2020}. Correspondingly, we randomly shuffle 50\% of the input sequences\footnote{Roughly 25\% of all sequences comprise only a single MRI screening. Therefore, we actually randomly shuffle 66\% of sequences considering sequences with a single observation as negative instances.} and task the model to differentiate shuffled from non-shuffled sequences. During the SOP pre-training, we keep the parameters of the CNN feature extractor fixed and initialize them using the previously pre-trained weights. We apply the sigmoid function to the output of the model's pooling layer and learn to identify shuffled sequences by minimizing the binary cross-entropy loss w.r.t. the parameters $\theta_{G^\prime} = \theta_G \cup \theta_\phi$ of the Transformer encoder $G$ and the pooling layer $\phi$:
\begin{equation}
    \min_{\theta_{G^\prime}} - \sum_{e \in \textbf{E}^{(p,t)}} t_{G^\prime}(e) \text{log}(p_{G^\prime}(e)),
\end{equation}

\noindent where $p_{G^\prime}(e)$ is the predicted probability of shuffling and $t_{G^\prime}(e)$ is the true value. Note that during pre-training of the Transformer encoder we do not randomly crop multiple views from the same image but instead only crop a single view of shape $s = 72^3$ from the original MRI. Likewise, we concatenate the different series in the sequence of MRI scans to a multi-channel volume instead of randomly selecting a single one. Provided a sequence of DW-MRI's with diffusion coefficients $b \in \{0, 150, 400, 800\}$, the input $i^{(p,t)}$ to the CNN backbone $G$ is thus a four rather than a one channel volume. 

\subsection{Fine-Tuning}
\label{sec:finetuning}
After pre-training, we initialize the parameters of HCCNet using the previously pre-trained weights. Additionally, we append the final pooling layer as discussed in \ref{sec:architecture} and fine-tune the model on our initial task by minimizing the class-balanced binary cross-entropy w.r.t. all parameters $\theta_{\text{HCCNet}}$: 
\begin{equation}
    \min_{\theta_{\text{HCCNet}}} - \omega \sum_{e \in \textbf{E}_{(p,t)}} \textbf{y}^p \text{log}(\hat{\textbf{y}}^p).
\end{equation}

\noindent where $\omega$ is the weighting factor for the positive class, $\hat{\textbf{y}}_p$ is the predicted probability of HCC development and $\textbf{y}_p$ is the observed development in patient $p$.

\section{Experiments}
\label{sec:experiments}

\subsection{Data Acquisition and Split}
\label{sec:data}
We evaluate HCCNet on data that is retrospectively collected from a cohort of liver cirrhosis patients who underwent repeated abdominal MRI screening for HCC surveillance between March 2011 and September 2022 at Medisch Spectrum Twente (MST) in Enschede, the Netherlands. Patients were followed up over 1 month to 11 years with screening intervals varying between 1 month and 6.5 years. In total, 963 MRI examinations were included in the study with longitudinal scans comprising contrast enhanced and non-contrast enhanced T1-, T2- and diffusion-weighted images which were acquired in axial orientation and registered using a Philips Ingenia Elition 1.5- and 3-Tesla scanner respectively. Prior to pre-training and fine-tuning, MR images were anonymized and resampled to 1.5x1.5x1.5mm voxel spacing. 

As described in section \ref{sec:problem_formulation}, we include all available data in superset $D_p$ during pre-training and subset the data for fine-tuning. During fine-tuning we additionally exclude all patients without a definitive diagnosis of HCC yielding a total of 243 patients of which 37 develop HCC (i.e., 703 and 101 total MRI examinations respectively). Furthermore, we randomly split superset $D_p$ in a development $D^{\text{dev}}_p$ and a test set $D^{\text{test}}_p$ using a 75:25 split. To avoid any information leakage between pre-training and fine-tuning, we only fine-tune HCCNet on longitudinal MRIs that are also included in superset $D^{\text{dev}}_p$. 


\subsection{Implementation Details and Baseline}
\label{sec:implementation}
\textbf{Pre-Training.} As aforementioned, we pre-train both the CNN backbone and the Transformer encoder on development superset $D_p^{\text{dev}}$. We train the CNN over 32000 training steps using the AdamW optimizer \citep{loshchilovDecoupledWeightDecay2019} with an effective batch size of $b = 128$. During the first 5\% of training steps (i.e., 1600 warm-up steps), we linearly increase the learning rate $\eta$ to its base value which is determined using the following square-root scaling rule \citep{granziolLearningRatesFunction}: $\eta = \alpha \frac{\sqrt{b}}{\sqrt{128}}$, where $b$ is the effective batch size and $\alpha$ is a constant scaling factor that is chosen according to $b$. After reaching its maximum value, we decay the learning rate using a cosine schedule. We employ the same cosine decay without warm-up for the weight decay $\lambda$ but only apply it to the weights of the parameters excluding normalization layers \citep{liUnderstandingDisharmonyWeight2020}. Contrary to the original DINO implementation, we also limit the temperature of the teacher network to a constant $\tau_{F_t} = 0.04$ and increase the initial exponential moving average momentum to $m = 0.9995$.

While MRI scans are assumed to be i.i.d. during CNN pre-training, pre-training of the Transformer encoder considers the within-person dependencies across time points. Hence, to mitigate the risk of overfitting due to the reduced number of independent samples, we train the Transformer encoder using an abbreviated pre-training protocol of just 8000 training steps with a learning rate warm-up of 5\% or 10\% for the smaller and bigger model variants respectively and a reduced effective batch size of $b = 32$. We set the maximal sequence length to $s = 8$, apply dropout with probability $p = 0.2$, and employ the same optimizer, learning rate scaling and cosine schedule as during CNN pre-training.

\textbf{Fine-Tuning.} During fine-tuning, we update HCCNet on the development subset $V_p^{\text{dev}}$. We fine-tune the model over 400 training steps with 5\% warm-up using the same optimizer, sequence length, and learning rate schedule as during pre-training of the Transformer encoder but decrease the weight decay to a constant $\lambda = 1e^{-5}$. For the individual MRI sequences, we choose the optimal batch size from $b \in \{16, 32, 64\}$ using 5-fold cross-validation with no further hyperparameter tuning and instead retrain the model on the full development subset $V_p^{\text{dev}}$ over 10 runs with different random seeds using the same default hyper-parameter settings as during Transformer encoder pre-training. 

\textbf{Baseline.} To contrast the proposed modeling approach, we establish a baseline by training HCCNet from scratch using randomly initialized parameters sampled from $\mathcal{N}(0, 0.02)$. To provide an unbiased comparison, we equally train the baseline on development subset $V^{(p,t)}_\text{dev}$ over 10 runs with different random seeds. We employ the same optimizer, learning rate schedule, number of training steps, batch size, and further hyper-parameter settings as during fine-tuning but slightly increase the learning rate warm-up (i.e., from 20 to 40 training steps) to ensure a more stable convergence. A full list of hyperparameter setting is presented in table \ref{tab:hyperparams} in \ref{sec:appendix_hyperparam}. 

\subsection{Evaluation Protocol}
\label{sec:evaluation}
After fine-tuning, we evaluate HCCNet's baseline and the fine-tuned predictive capability of HCC on the test set $V_{\text{test}}^{(p,t)}$ by reporting the average performance over the different model runs on a central crop. For evaluation, we employ the following performance metrics: accuracy, precision, recall, F1-score, area under the receiver operating characteristic (AUROC), and area under the precision-recall curve (AUPRC). For the comparison across model variations, we also explore the robustness of the model by comparing the mean absolute error (MAE) formulated as:
\begin{equation}
    \text{MAE} = \frac{1}{n}\sum_{i=1}^n | \text{gain}^\prime_{\text{obs},i} - \text{gain}_{\text{exp},i} |,
\end{equation}

\noindent where $i$ represents the individual model run, ranging from $i$ to $n$, $\text{gain}^{\text{exp},i}$ denotes the expected cumulative gain equal to the identity line, and $\text{gain}^\prime_{\text{obs},i}$ represents the normalized observed cumulative gain, which is computed over $\text{gain}^\prime_{\text{obs},i} = \frac{\sum_{k=1}^i m_k}{i}$, for $i = 1, 2, ..., n$, where $m_k$ denotes the metric over which the cumulative gain is computed. Likewise, for the comparison against the randomly initialized baseline, we evaluate the models' respective calibration considering the expected (ECE), and maximum calibration error (MCE) as well as the Brier score.

\subsection{Main Results}
\label{sec:results}
\begin{table}
    \centering
    \caption{Fine-Tuned Model Results for the Prediction of HCC at the Next Observation}
    \label{tab:finetuned_results}
    \resizebox{\textwidth}{!}{%
    \begin{threeparttable}
    \begin{tabular}{lccccccc}
        \toprule
        & & \multicolumn{6}{c}{Performance Metrics} \\
        \cmidrule(lr){3-8}
        & & Accuracy & Precision & Recall & F1-Score & AUPRC & AUROC \\ 
        \midrule
        DW-MRI & HCCNet-F & 0.874 $\pm$ 0.040 & \underline{0.578 $\pm$ 0.141} & 0.787 $\pm$ 0.159 & 0.639 $\pm$ 0.047 & \underline{0.690 $\pm$ 0.030} & 0.916 $\pm$ 0.011 \\
        & HCCNet-P & \textbf{0.914 $\pm$ 0.017} & \textbf{0.668 $\pm$ 0.087} & 0.812 $\pm$ 0.084 & \textbf{0.724 $\pm$ 0.031} & \textbf{0.744 $\pm$ 0.051} & \textbf{0.936 $\pm$ 0.013} \\
        & HCCNet-N & 0.881 $\pm$ 0.020 & 0.547 $\pm$ 0.047 & \textbf{0.900 $\pm$ 0.050} & 0.678 $\pm$ 0.032 & 0.689 $\pm$ 0.054 & \underline{0.930 $\pm$ 0.013} \\
        & HCCNet-T & \underline{0.888 $\pm$ 0.018} & 0.565 $\pm$ 0.049 & \underline{0.887 $\pm$ 0.118} & \underline{0.685 $\pm$ 0.044} & 0.624 $\pm$ 0.064 & 0.928 $\pm$ 0.012 \\
        T1 DCE-MRI & HCCNet-F & 0.868 $\pm$ 0.024 & 0.537 $\pm$ 0.077 & 0.625 $\pm$ 0.097 & 0.571 $\pm$ 0.060 & 0.435 $\pm$ 0.060 & 0.791 $\pm$ 0.036 \\
        & HCCNet-P & 0.826 $\pm$ 0.059 & 0.472 $\pm$ 0.134 & 0.713 $\pm$ 0.126 & 0.544 $\pm$ 0.064 & 0.449 $\pm$ 0.072 & 0.779 $\pm$ 0.030 \\
        & HCCNet-N & 0.800 $\pm$ 0.094 & 0.441 $\pm$ 0.132 & 0.675 $\pm$ 0.150 & 0.504 $\pm$ 0.066 & 0.386 $\pm$ 0.069 & 0.728 $\pm$ 0.048 \\
        & HCCNet-T & 0.881 $\pm$ 0.025 & 0.604 $\pm$ 0.149 & 0.600 $\pm$ 0.094 & 0.587 $\pm$ 0.058 & 0.535 $\pm$ 0.074 & 0.781 $\pm$ 0.046 \\
        T1 IOP \& T2-MRI & HCCNet-F & 0.665 $\pm$ 0.101 & 0.257 $\pm$ 0.064 & 0.625 $\pm$ 0.137 & 0.350 $\pm$ 0.046 & 0.217 $\pm$ 0.029 & 0.611 $\pm$ 0.043 \\
        & HCCNet-P & 0.746 $\pm$ 0.070 & 0.317 $\pm$ 0.067 & 0.613 $\pm$ 0.067 & 0.412 $\pm$ 0.058 & 0.281 $\pm$ 0.094 & 0.672 $\pm$ 0.045 \\
        & HCCNet-N & 0.802 $\pm$ 0.045 & 0.396 $\pm$ 0.084 & 0.600 $\pm$ 0.094 & 0.462 $\pm$ 0.031 & 0.338 $\pm$ 0.059 & 0.708 $\pm$ 0.022 \\
        & HCCNet-T & 0.732 $\pm$ 0.206 & 0.353 $\pm$ 0.101 & 0.562 $\pm$ 0.170 & 0.405 $\pm$ 0.064 & 0.266 $\pm$ 0.061 & 0.664 $\pm$ 0.063 \\
        \bottomrule
    \end{tabular}
    \begin{tablenotes}
        \Large
        \item \textit{Note:} Performance metrics range from 0 to 1 or from 0.50 to 1 and from 0.14 to 1 for the AUROC and the AUPRC respectively. All metrics are presented using the mean and standard deviation across the individual 10 model runs. The best model results are emphasised in boldface. Second best results are underlined. DW-MRI comprises diffusion MRI scans with diffusion coefficient $b \in \{0, 150, 400, 800\}$. T1 DCE-MRI contain a pre-contrast and three post-contrast (i.e., late aterial, portal venous, and delayed phase) series. T1 IOP \& T2-MRI comprise T1-weighted in- and out-of-phase MRI as well as T2-weighted MRI with short and long echo times.
    \end{tablenotes}
    \end{threeparttable}
    }
    \normalsize
\end{table}

We report HCCNet's fine-tuned performance (i) across imaging modalities, (ii) model variations, and (iii) against a randomly initialized baseline. Additional model results are presented in \ref{sec:appendix_results}.

\textbf{Comparison across Imaging Modalities.} Table \ref{tab:finetuned_results} presents fine-tuned model results for the prediction of HCC development at the next observation stratified by imaging modality and model architecture. HCCNet trained using diffusion weighted MR images thereby shows excellent predictive capability of tumor developement with an average $\text{AUPRC} = 0.687$ and $\text{AUROC} = 0.928$ across model architectures. Contrast enhanced T1 weighted and the combination of T1 weighted in- and out-of-phase and T2 weighted images, on the other hand, appear to display less expressive characteristics for the prediction of HCC development, albeit models trained using contrast enhanced T1-weighted images still maintain a significantly stronger average predictive capability than the combined T1 IOP and T2-weighted MRIs (i.e., AUPRC = 0.451 and AUROC = 0.770 versus AUPRC = 0.276 and AUROC = 0.664).

\begin{figure}[t!]
    \centering
    \caption{Cumulative Gain Across Different Model Runs}
    \includegraphics[width=0.99\linewidth]{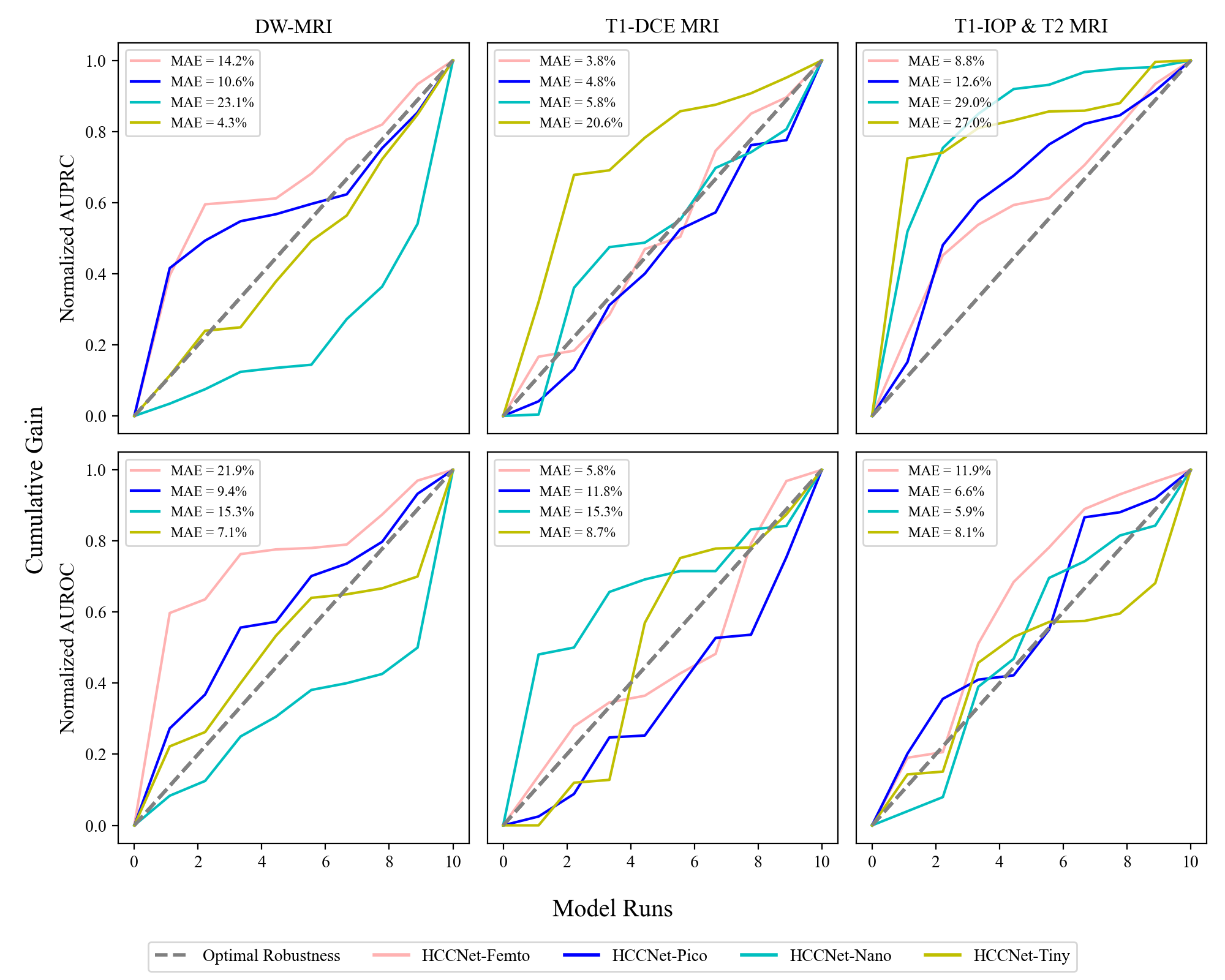}
    \label{fig:cumulative_gain}
    \hrule
    \begin{minipage}{0.99\textwidth}
        \vspace{1mm}
        \footnotesize
        \emph{Note}: Normalized AUROC and AUPRC values are presented for multiple runs of different HCCNet model variants (femto, pico, nano, tiny) across three MRI types (DW-MRI, T1-DCE MRI, T1-IOP \& T2 MRI). Each colored line represents the cumulative gain in normalized performance. Optimal robustness is presented by the gray dashed line, indicating no sensitivity to random variations. The Mean Absolute Error (MAE) quantifies the deviation of each model’s performance from this optimal line averaged over 10 runs with different random seeds. Generally, lower MAE values indicate greater robustness to random variations, while higher MAE values signify greater sensitivity to individual model runs.
    \end{minipage}
\end{figure}


\textbf{Comparison across Model Variations.} In addition to the varying image modalities, we explore the impact of varying model complexity on HCC risk prediction. As such, Figure \ref{fig:cumulative_gain} plots the cumulative gain in normalized AUPRC and AUROC against the test results obtained over the 10 individual model runs and ordered by ascending cumulative gain. For the models fine-tuned on diffusion weighted MR images, the cumulative gain plots indicate some sensitivity to model runs. In fact, they imply that HCCNet-nano's high average AUPRC and AUROC may to some extent be attributed to a few individual model runs, where the model performs exceptionally well, while HCCNet-femto's performance may likely be diminished by a few runs, where the model performs worse than expected. Generally, however, models closely follow the diagonal line and expect for HCCNet-nano fine-tuned on DW-MRIs no model falls consistently below the diagonal line. 

Across imaging modalities, the models are thus robust and not overly sensitive to seed variations. Likewise, the plots reveal no consistent differences across model architecture, as no model variant is persistently deviating from the optimal diagonal line. This corresponds with the results presented in Table \ref{tab:finetuned_results}, which indicate that compared to the differences between imaging modalities model size only negligibly affects downstream model performance. At least on the data evaluated in this study, variations in model complexity do thus not considerably influence the quality of the models' predictions nor their robustness.

\begin{table}
    \centering
    \caption{Baseline vs. Fine-Tuned Model Results}
    \label{tab:results_base_pre}
    \resizebox{\textwidth}{!}{%
    \begin{threeparttable}
    \begin{tabular}{llccS[table-format=1.1]ccS[table-format=1.1]}
        \toprule
        & & \multicolumn{3}{c}{AUPRC} & \multicolumn{3}{c}{AUROC} \\
        \cmidrule(lr){3-8}
        & & Baseline & Fine-Tuned & {Rel. Change} & Baseline & Fine-Tuned & {Rel. Change} \\ 
        \midrule
        DW-MRI & HCCNet-F & 0.286 $\pm$ 0.041 & \underline{0.690 $\pm$ 0.031} & \underline{141.5\%} & 0.706 $\pm$ 0.025 & 0.916 $\pm$ 0.011 & 29.8\% \\
        & HCCNet-P & 0.309 $\pm$ 0.063 & \textbf{0.744 $\pm$ 0.054} & 141.1\% & 0.715 $\pm$ 0.017 & \textbf{0.936 $\pm$ 0.013} & \textbf{30.9\%} \\
        & HCCNet-N & 0.269 $\pm$ 0.057 & 0.689 $\pm$ 0.056 & \textbf{156.1\%} & 0.712 $\pm$ 0.023 & \underline{0.930 $\pm$ 0.014} & \underline{30.7\%} \\
        & HCCNet-T & 0.311 $\pm$ 0.047 & 0.624 $\pm$ 0.068 & 100.5\% & 0.716 $\pm$ 0.037 & 0.928 $\pm$ 0.013 & 29.5\% \\
        T1 DCE-MRI & HCCNet-F & 0.389 $\pm$ 0.083 & 0.436 $\pm$ 0.063 & 12.0\% & \underline{0.755 $\pm$ 0.051} & 0.790 $\pm$ 0.039 & 4.7\% \\
        & HCCNet-P & \underline{0.402 $\pm$ 0.054} & 0.450 $\pm$ 0.076 & 11.9\% & \textbf{0.773 $\pm$ 0.027} & 0.777 $\pm$ 0.031 & 0.4\% \\
        & HCCNet-N & 0.340 $\pm$ 0.064 & 0.388 $\pm$ 0.073 & 14.0\% & 0.746 $\pm$ 0.044 & 0.727 $\pm$ 0.050 & -2.6\% \\
        & HCCNet-T & 0.361 $\pm$ 0.049 & 0.535 $\pm$ 0.079 & 48.4\% & 0.751 $\pm$ 0.043 & 0.779 $\pm$ 0.048 & 3.6\% \\
        T1 IOP \& T2-MRI & HCCNet-F & \textbf{0.427 $\pm$ 0.133} & 0.223 $\pm$ 0.031 & -47.7\% & 0.672 $\pm$ 0.050 & 0.615 $\pm$ 0.046 & -8.5\% \\
        & HCCNet-P & 0.349 $\pm$ 0.088 & 0.282 $\pm$ 0.099 & -19.4\% & 0.675 $\pm$ 0.051 & 0.666 $\pm$ 0.049 & -1.2\% \\
        & HCCNet-N & 0.319 $\pm$ 0.049 & 0.338 $\pm$ 0.063 & 5.9\% & 0.642 $\pm$ 0.033 & 0.703 $\pm$ 0.023 & 9.6\% \\
        & HCCNet-T & 0.298 $\pm$ 0.069 & 0.268 $\pm$ 0.065 & -10.0\% & 0.632 $\pm$ 0.043 & 0.665 $\pm$ 0.066 & 5.3\% \\
        \bottomrule
    \end{tabular}
    \begin{tablenotes}
        \Large
        \item \textit{Note:} AUPRC and the AUROC values range from 0.14 and 0.5 to 1 respectively. All metrics are presented using the mean and standard deviation across the individual 10 model runs. The best results are emphasised in boldface. Second best results are underlined. DW-MRI comprises diffusion MRI scans with diffusion coefficient $b \in \{0, 150, 400, 800\}$. T1 DCE-MRI comprises a pre-contrast and three post-contrast (i.e., late aterial, portal venous, and delayed phase) series, while T1 IOP \& T2-MRI comprise T1-weighted in- and out-of-phase MRI as well as T2-weighted MRI with short and long echo times.
    \end{tablenotes}
    \end{threeparttable}
    }
    \normalsize
\end{table}

\textbf{Comparison against Baseline.} Finally, we evaluate the impact of our proposed pre-training approach on model performance. Table \ref{tab:results_base_pre} presents baseline and fine-tuned model results stratified by image modality and model variation. Likewise, Table \ref{tab:results_base} in \ref{sec:appendix_baseline} presents extended baseline model results. Across imaging modalities, we notice very high variability in the success of our proposed approach. That is, although baseline model results vary only gradually across image modalities, diffusion weighted MR images display an average improvement of 134.8\% in AUPRC and 30.2\% in AUROC after fine-tuning, while dynamic contrast enhanced T1-weighted images only show moderate improvements (i.e., 21.6\% in AUPRC and 1.5\% in AUROC), and the combination of T1-weighted in- and out-of-phase and T2-weighted MRIs even displays an average reduction of -17.8\% in AUPRC and only a small improvement of 1.3\% in AUROC after fine-tuning. 

\begin{figure}[t!]
    \centering
    \caption{Baseline vs. Fine-Tuned Models' Reliability Diagrams}
    \includegraphics[width=0.99\linewidth]{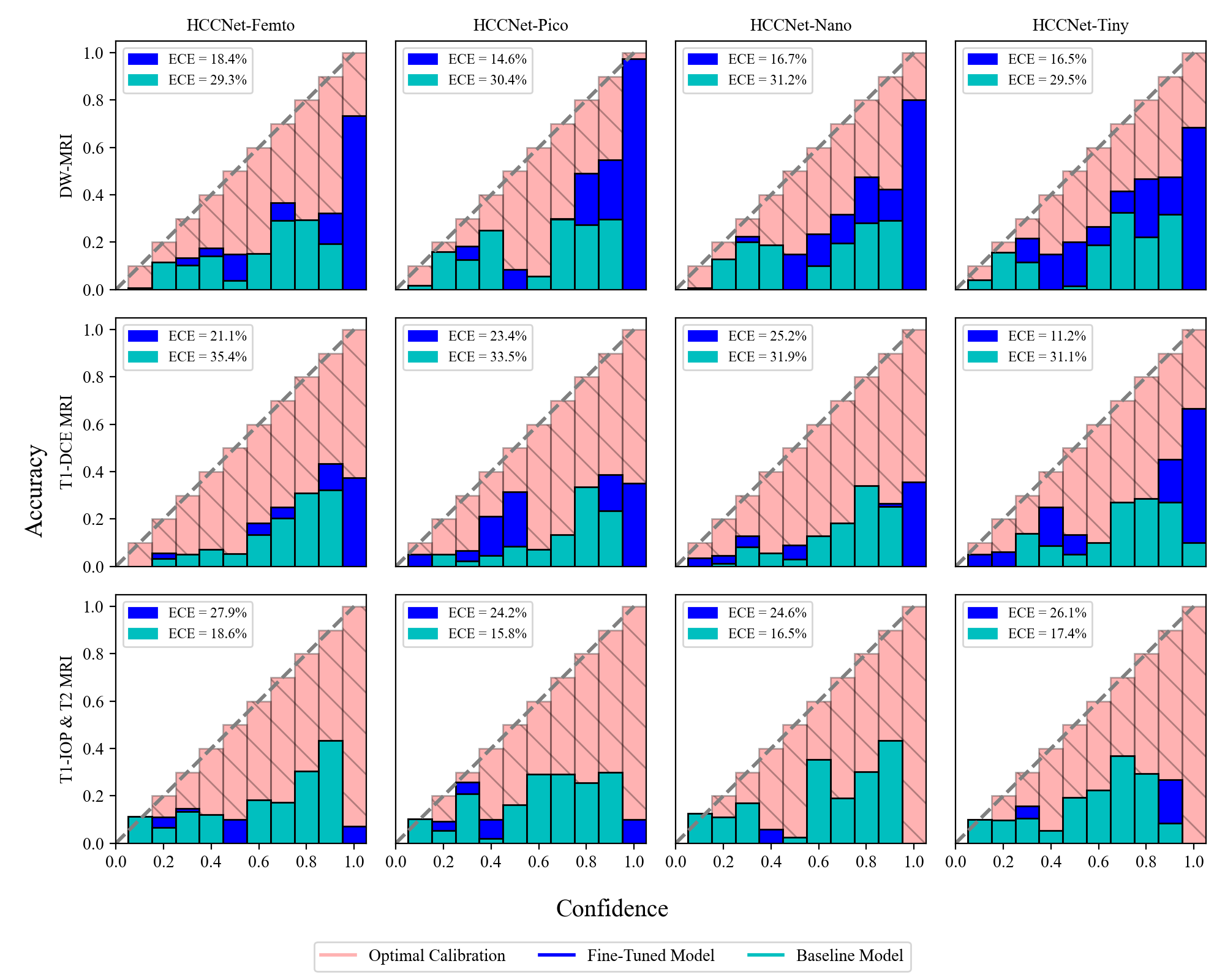}
    \label{fig:calibration}
    \hrule
    \begin{minipage}{0.99\textwidth}
        \vspace{1mm}
        \footnotesize
        \emph{Note}: Calibration plots of HCCNet model variants (femto, pico, nano, tiny) are shown across three MRI modalities (DW-MRI, T1-DCE MRI, T1-IOP \& T2 MRI) to evaluate the alignment between predicted confidence and actual accuracy. The x-axis represents confidence levels (0.0 to 1.0), while the y-axis shows accuracy within each confidence bin. Optimal calibration, where predicted confidence matches actual accuracy, is depicted by the pink-striped region. Blue bars show the calibration of fine-tuned models, and green bars represent baseline models. Expected Calibration Error (ECE), shown in the legend as a percentage, quantifies the average mismatch between confidence and accuracy across bins; lower ECE indicates better calibration.
    \end{minipage}
\end{figure}


To further investigate the apparent disparity in pre-training success, we compare the reliability of the baseline models' predictions to the reliability of the fine-tuned models' predictions. Accordingly, Figure \ref{fig:calibration} presents reliability diagrams across imaging modalities and model variants. Extended reliability metrics are presented in Table \ref{tab:reliability_results} in \ref{sec:appendix_reliability}. For the fine-tuned models, the plots largely align with the results presented in Table \ref{tab:results_base_pre}. Excluding HCCNet-Tiny fine-tuned using dynamic contrast enhanced T1 MRIs, Figure \ref{fig:calibration} shows consistently lower expected calibration errors among models fine-tuned using diffusion weighted MR images and higher expected calibration errors among models fine-tuned on T1 in- and out-of-phase and T2-weighted MRIs. For the randomly initialized baseline, however, the expected calibration error does not display the same negative relationship with model performance. Here, baseline models trained using T1-IOP and T2-weighted MR images consistently show the lowest expected calibration error, while models trained using dynamic contrast enhanced T1-weighted MRIs display the highest expected calibration error. 

More interestingly, the comparison between baseline and fine-tuned models' reliability diagrams further indicates that despite only modest improvements in AUROC and AUPRC, models fine-tuned using T1-DCE MR images display an average reduction in expected calibration error of 38.7\% compared to the models trained with randomly initialized parameters. Although the proposed self-supervised pre-training approach does thus not always yield greatly improved performance metrics, even small gains lead to more reliable and trustworthy confidence scores, making the models predicted scores of tumor development more dependable for probabilistic interpretation.

\section{Discussion}
\label{sec:discussion}
Despite being long established as the gold standard in many branches of oncology, deep learning applications to the study of longitudinal medical images remain notoriously understudied. Therefore, this study developed HCCNet—a spatio-temporal neural network that utilizes a 3D ConvNeXt backbone combined with a Transformer encoder—to predict future cancer development based on past MRI examinations. To facilitate HCCNet's capability to capture disease progression over time, a step-wise model training approach was adopted. We first pre-trained HCCNet's CNN backbone and Transformer encoder using a custom self-supervised pre-training framework tailored for longitudinal medical images and then fine-tuned the model on our downstream task, initializing HCCNet's parameters with the pre-trained weights.

Overall, fine-tuned model results varied substantially across imaging modalities. Among them, diffusion weighted MR images (DW-MRIs) displayed the highest predictive performance, achieving an average AUC-PR of 0.687 and AUC-ROC of 0.928. In comparison, contrast-enhanced T1-weighted MRIs (T1 DCE-MRIs) showed moderate predictive power, while combined T1 in- and out-of-phase and T2-weighted MRIs exhibited the lowest accuracy. Moreover, our proposed pre-training approach improved model performance by up to 156.1\% for models fine-tuned using diffusion-weighted MR images but showed only slight or even reduced performance metrics for models fine-tuned on dynamic contrast-enhanced T1-weighted and T1-IOP and T2-weighted MR images.

Inspection of the models' reliability diagrams further demonstrated that the proposed self-supervised pre-training approach aligned the models closer to the target distribution, reducing the extent to which calibration errors are propagated and amplified during fine-tuning. Thus, even for imaging modalities where the pre-training approach only yielded moderate improvements, the predicted risks were more reliable, demonstrating the value of our proposed pre-training framework for applications in the medical domain.


Despite these successes, several challenges and limitations were identified. The variability in pre-training success across different MRI modalities suggests that further optimization of the pre-training process is needed to ensure consistent performance across various imaging types. Additionally, while HCCNet demonstrates strong predictive capabilities for HCC development, its application to other diseases will require careful consideration of the specific characteristics and progression patterns of those diseases.

Another important point is the potential for integrating additional clinical data, such as electronic health records (EHR) or molecular biomarkers, into the HCCNet framework. Incorporating these data sources could further enhance the model’s predictive power and provide a more comprehensive understanding of disease progression. Furthermore, the study’s focus on a single cohort from a specific clinical setting may limit the generalizability of the findings, highlighting the need for external validation in diverse patient populations.

\section{Conclusion}
\label{sec:conclusion}
In this work, we presented a framework for longitudinal medical image analysis and its application to the study of HCC early detection. The methodology is versatile and offers a framework for other longitudinal medical imaging applications where patient monitoring can lead to better outcome. Yet, each condition presents its own unique challenges in terms of disease progression and imaging characteristics, which will require further adaptation of the HCCNet framework. Additionally, exploring the integration of multi-modal data, including EHR, genomic information, and other imaging modalities, could provide a more holistic view of patient health and improve predictive accuracy, while further external validation is needed to assess the generalizability of the study's findings. In the future, we therefore plan to validate our findings on an external validation cohort and aim to incorporate multi-modal data to further improve upon our results. Likewise, future research is needed to investigate the versatility of the proposed methodology and its application to diseases beyond HCC.
\section*{Acknowledgment}
This research was performed within the project entitled \textit{CHARISMA: Predictie van hepatocellulair carcinoom op basis van MRI-radiomics
Primair leverkanker (hepatocellulair carcinoma, HCC)}, which is funded by  \textit{Pioneers in Health Care} (PIHC) Innovation Fund, Twente region, The Netherlands. We also acknowledge the use of OpenAI's ChatGPT-4 for assistance in refining the English language of this manuscript.

\clearpage
\appendix

\section{Implementation Details}
\subsection{3D ConvNeXt Block Design}
\label{sec:appendix_conv_layer}
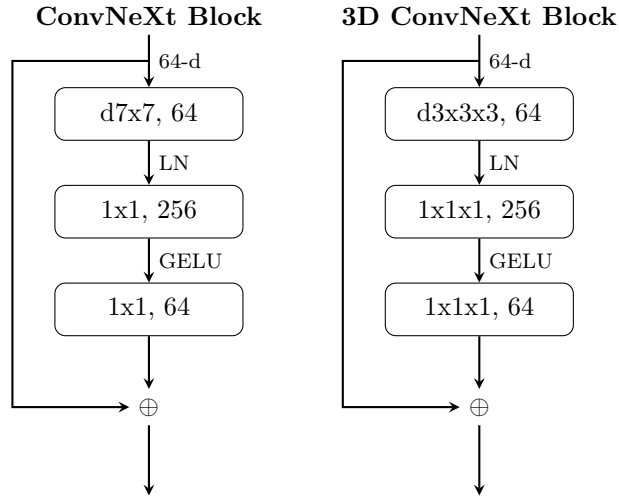
\begin{figure}[H]
    \centering
    \caption{Original and 3D ConvNeXt Block Designs}
    \begin{tikzpicture}[scale=0.33, node distance=1.3cm]

        \node (header) [font=\bfseries] {ConvNeXt Block};
        \node (conv1) [element, below of=header] {d7x7, 64};
        \node (conv2) [element, below of=conv1] {1x1, 256};
        \node (conv3) [element, below of=conv2] {1x1, 64};
        \node (concat) [below of=conv3] {$\oplus$};
        \node (output) [below of=concat] {};

        \draw [arrow] (header) -- node[right, font=\footnotesize] {64-d} (conv1);
        \draw [arrow] (conv1) -- node[right, font=\footnotesize] {LN} (conv2);
        \draw [arrow] (conv2) -- node[right, font=\footnotesize] {GELU} (conv3);
        \draw [arrow] (conv3) -- (concat);
        \draw [arrow] (concat) -- (output);
        \draw [arrow] ($(header.south)!0.5!(conv1.north)$) -| ++(-5.5,0) -- ++(0,-6.2) |- (concat);
        
    \end{tikzpicture}
    \qquad
    \begin{tikzpicture}[scale=0.33, node distance=1.3cm]

        \node (header) [font=\bfseries] {3D ConvNeXt Block};
        \node (conv1) [element, below of=header] {d3x3x3, 64};
        \node (conv2) [element, below of=conv1] {1x1x1, 256};
        \node (conv3) [element, below of=conv2] {1x1x1, 64};
        \node (concat) [below of=conv3] {$\oplus$};
        \node (output) [below of=concat] {};

        \draw [arrow] (header) -- node[right, font=\footnotesize] {64-d} (conv1);
        \draw [arrow] (conv1) -- node[right, font=\footnotesize] {LN} (conv2);
        \draw [arrow] (conv2) -- node[right, font=\footnotesize] {GELU} (conv3);
        \draw [arrow] (conv3) -- (concat);
        \draw [arrow] (concat) -- (output);
        \draw [arrow] ($(header.south)!0.5!(conv1.north)$) -| ++(-5.5,0) -- ++(0,-6.2) |- (concat);
        
    \end{tikzpicture}
    \label{fig:block_design}
    \hrule
    \begin{minipage}{0.99\textwidth}
        \vspace{1mm}
        \footnotesize
        \emph{Note}: The figure illustrates the structure of a ConvNeXt block with an input feature map of dimensionality $d = 64$. The input is passed through a series of convolutional layers, where the leading d denotes the use of depthwise convolutions, before being merged with the block's initial input through a residual connection. The figure's left hand side represents the original implementation in \citep{liuConvNet2020s2022}, while the right illustrates the study's 3D adaptation.
    \end{minipage}
\end{figure}

Figure \ref{fig:block_design} illustrates the structure of the original ConvNeXt block as implemented in \citep{liuConvNet2020s2022} as well as the 3-dimensional adaptation proposed in this study. Compared to the original design, we inflate the 2D depthwise and pointwise convolutional layers to 3 dimensions and reduce the depthwise convolutional layer's proposed kernel size from $k = 7$ to $k = 3$, leaving the remaining block design unchanged.

\subsection{Data Augmentation Strategy}
\label{sec:appendix_augmentation}
During pre-training of the CNN feature extractor, we randomly crop two global views of shape $s = 72^3$ and two local view of shape $s = 48^3$ from the image, randomly selecting a single series out of the sequence of provided MRI's for each of the cropped views. Contrarily, during pre-training of the Transformer encoder, we randomly crop a multi-channel sub-volume of shape $s = 72^3$ from the image, concatenating all available series from the sequence of MRI's to a single 4D image. Apart from the alternating cropping approach, further augmentations remain consistent across pre-training. Specifically, we randomly shift and scale the intensity by a constant $v$ using probability $p = 1.0$, apply random 90 degree rotations with probability $p = 0.5$, and randomly flip the cropped view along the three axes with probability $p = 0.5$. During fine-tuning as well as for the baseline model, we utilize the same data augmentation strategy as during pre-training of the Transformer encoder. 

\subsection{Hyperparameter Settings}
\label{sec:appendix_hyperparam}
\begin{table}[H]
    \centering
    \caption{Hyperparameter Settings}
    \label{tab:hyperparams}
    \resizebox{\textwidth}{!}{%
    \begin{threeparttable}
    \begin{tabular}{lcccc}
        \toprule
        & \multicolumn{2}{c}{Pre-Training} & \multicolumn{1}{c}{Fine-Tuning} & \multicolumn{1}{c}{Baseline} \\
        \cmidrule(lr){2-3}
        & Backbone & Encoder & & \\ 
        \midrule
        Optimizer & AdamW & AdamW & AdamW & AdamW \\
        Weight Initialization & $\mathcal{N}(0, 0.02)$ & Partially Pre-Trained & Pre-Trained & $\mathcal{N}(0, 0.02)$ \\
        Training Steps & 32000 & 8000 & 400 & 400 \\
        Warm-up Steps & 1600 & [400, 800] & 20 & 40 \\
        Effective Batch Size & 128 & 32 & [16, 32, 64] & [16, 32, 64] \\
        Sequence Length & - & 8 & 8 & 8 \\
        Base Learning Rate & 4e-4 & 1e-4 & [5e-5, 1e-4, 2e-4] & [5e-5, 1e-4, 2e-4]  \\
        Minimum Learning Rate & 1e-6 & 1e-6 & 1e-6 & 1e-6 \\
        Weight Decay & 5e-2 & 5e-2 & 1e-5 & 1e-5 \\
        Maximum Weight Decay & 5e-1 & 5e-1 & 1e-5 & 1e-5 \\
        Label Smoothing & - & - & 0.1 & 0.1 \\
        Dropout & - & 0.2 & 0.2 & 0.2 \\
        Gradient Clipping & 1.0 & 1.0 & 3.0 & 3.0 \\
        Teacher Temperature & 0.04 & - & - & - \\
        Student Temperature & 0.1 & - & - & - \\
        Projection Head Dimensionality & 1024 & - & - & - \\
        Exponential Moving Average & 0.9995 & - & - & - \\
        Random Flip and Rotation & 0.5 & 0.5 & 0.5 & 0.5 \\
        Random Intensity Scaling and Shifting & 1.0 & 1.0 & 1.0 & 1.0 \\
        \bottomrule
    \end{tabular}
    \begin{tablenotes}
        \large
        \item \textit{Note:} During pre-training of the Transformer encoder, model parameters of the CNN backbone are initialized using pre-trained weights, whereas parameters of the transformer encoder are randomly initialized. The learning rate is linearly increased over 400 training steps for model variations HCCNet-F and -P and over 800 training steps for HCCNet-N and -T. For all phases of training, the base learning rate $\eta$ is computed as a function of the batch size $b$ using the following formula: $\eta = \alpha \frac{\sqrt{b}}{\sqrt{128}}$, where $\alpha$ is a common scaling factor that is chosen according to $b$.
    \end{tablenotes}
    \end{threeparttable}
    }
    \normalsize
\end{table}

Table \ref{tab:hyperparams} presents detailed hyperparameter settings for pre-training, fine-tuning, and the baseline respectively. Note that during CNN-backbone pre-training we slightly deviate from the hyperparameter settings proposed in \citep{caronEmergingPropertiesSelfSupervised2021}. This is due to convergence issues that we noted during initial model runs. Consequently, we drastically reduced the projection head’s output dimensionality to 1024 classes and reduced the temperature of the teacher network to a constant $\tau_{F_t} = 0.04$. For the other parameters, we followed the recommendations in \citep{caronEmergingPropertiesSelfSupervised2021} and increased the initial momentum of the exponential moving average update to $m = 0.9995$ given our reduced batch size of $B = 128$.

\section{Additional Results}
\label{sec:appendix_results}
\subsection{Extented Baseline Model Results}
\label{sec:appendix_baseline}
\begin{table}[H]
    \centering
    \caption{Baseline Model Results for the Prediction of HCC at the Next Observation}
    \label{tab:results_base}
    \resizebox{\textwidth}{!}{%
    \begin{threeparttable}
    \begin{tabular}{lccccccc}
        \toprule
        & & \multicolumn{6}{c}{Performance Metrics} \\
        \cmidrule(lr){3-8}
        & & Accuracy & Precision & Recall & F1-Score & AUPRC & AUROC \\ 
        \midrule
        DW-MRI & HCCNet-F & 0.721 $\pm$ 0.067 & 0.296 $\pm$ 0.045 & 0.675 $\pm$ 0.139 & 0.403 $\pm$ 0.022 & 0.286 $\pm$ 0.039 & 0.706 $\pm$ 0.024 \\
        & HCCNet-P & 0.741 $\pm$ 0.035 & 0.305 $\pm$ 0.024 & 0.662 $\pm$ 0.080 & 0.415 $\pm$ 0.023 & 0.309 $\pm$ 0.059 & 0.715 $\pm$ 0.016 \\
        & HCCNet-N & 0.717 $\pm$ 0.031 & 0.288 $\pm$ 0.022 & 0.700 $\pm$ 0.061 & 0.407 $\pm$ 0.021 & 0.269 $\pm$ 0.054 & 0.712 $\pm$ 0.022 \\
        & HCCNet-T & 0.747 $\pm$ 0.045 & 0.317 $\pm$ 0.046 & 0.662 $\pm$ 0.126 & 0.418 $\pm$ 0.025 & 0.311 $\pm$ 0.044 & 0.717 $\pm$ 0.035 \\
        T1 DCE-MRI & HCCNet-F & 0.768 $\pm$ 0.058 & 0.369 $\pm$ 0.093 & 0.725 $\pm$ 0.156 & 0.470 $\pm$ 0.060 & 0.388 $\pm$ 0.080 & \underline{0.756 $\pm$ 0.048} \\
        & HCCNet-P & 0.772 $\pm$ 0.055 & 0.378 $\pm$ 0.103 & \textbf{0.750 $\pm$ 0.112} & \textbf{0.486 $\pm$ 0.046} & \underline{0.400 $\pm$ 0.051} & \textbf{0.773 $\pm$ 0.026} \\
        & HCCNet-N & 0.775 $\pm$ 0.049 & 0.365 $\pm$ 0.053 & \underline{0.738 $\pm$ 0.163} & \underline{0.479 $\pm$ 0.057} & 0.339 $\pm$ 0.060 & 0.748 $\pm$ 0.043 \\
        & HCCNet-T & 0.784 $\pm$ 0.039 & 0.372 $\pm$ 0.051 & 0.700 $\pm$ 0.100 & 0.478 $\pm$ 0.024 & 0.359 $\pm$ 0.047 & 0.752 $\pm$ 0.041 \\
        T1 IOP \& T2-MRI & HCCNet-F & \underline{0.811 $\pm$ 0.079} & \underline{0.438 $\pm$ 0.152} & 0.525 $\pm$ 0.050 & 0.458 $\pm$ 0.081 & \textbf{0.426 $\pm$ 0.126} & 0.677 $\pm$ 0.046 \\
        & HCCNet-P & \textbf{0.828 $\pm$ 0.044} & \textbf{0.450 $\pm$ 0.135} & 0.487 $\pm$ 0.067 & 0.449 $\pm$ 0.045 & 0.349 $\pm$ 0.083 & 0.679 $\pm$ 0.049 \\
        & HCCNet-N & 0.782 $\pm$ 0.058 & 0.348 $\pm$ 0.070 & 0.525 $\pm$ 0.075 & 0.410 $\pm$ 0.044 & 0.319 $\pm$ 0.046 & 0.646 $\pm$ 0.032 \\
        & HCCNet-T & 0.786 $\pm$ 0.037 & 0.338 $\pm$ 0.069 & 0.500 $\pm$ 0.056 & 0.399 $\pm$ 0.051 & 0.298 $\pm$ 0.065 & 0.637 $\pm$ 0.039 \\
        \bottomrule
    \end{tabular}
    \begin{tablenotes}
        \Large
        \item \textit{Note:} AUPRC and the AUROC values range from 0.14 and 0.5 to 1 respectively. All metrics are presented using the mean and standard deviation across the individual 10 model runs. The best results are emphasised in boldface. Second best results are underlined. DW-MRI comprises diffusion MRI scans with diffusion coefficient $b \in \{0, 150, 400, 800\}$. T1 DCE-MRI comprises a pre-contrast and three post-contrast (i.e., late aterial, portal venous, and delayed phase) series.
    \end{tablenotes}
    \end{threeparttable}
    }
    \normalsize
\end{table}

Table \ref{tab:results_base} presents extended baseline model results for the next observation perdiction of HCC stratified by image modality and model variation. Across modalities, baseline models trained using dynamic contrast enhanced T1-weighted MRIs consistently display the highest detection rate of HCC as well as the greatest average discriminatory capability. Overall differences between imaging modalities are however far less pronounced than for the models intialized with SSL pre-trained weighted (s. Table \ref{tab:finetuned_results}) and the overall diagnostic performance remains generally poor.

\subsection{Reliability Analysis}
\label{sec:appendix_reliability}
\begin{table}[H]
    \centering
    \caption{Baseline vs. Fine-Tuned Reliability Scores}
    \label{tab:reliability_results}
    \resizebox{\textwidth}{!}{%
    \begin{threeparttable}
    \begin{tabular}{llcccccc}
        \toprule
        & & \multicolumn{3}{c}{Baseline} & \multicolumn{3}{c}{Fine-Tuned} \\
        \cmidrule(lr){3-8}
        & & ECE & MCE & Brier Score & ECE & MCE & Brier Score \\ 
        \midrule
        DW-MRI & HCCNet-F & 0.293 $\pm$ 0.029 & 0.685 $\pm$ 0.048 & 0.237 $\pm$ 0.016 & 0.184 $\pm$ 0.016 & 0.699 $\pm$ 0.097 & 0.130 $\pm$ 0.010 \\
        & HCCNet-P & 0.304 $\pm$ 0.028 & 0.670 $\pm$ 0.090 & 0.245 $\pm$ 0.021 & \underline{0.146 $\pm$ 0.020} & \textbf{0.569 $\pm$ 0.140} & \textbf{0.091 $\pm$ 0.008} \\
        & HCCNet-N & 0.312 $\pm$ 0.040 & 0.737 $\pm$ 0.150 & 0.248 $\pm$ 0.024 & 0.167 $\pm$ 0.024 & \underline{0.592 $\pm$ 0.153} & \underline{0.102 $\pm$ 0.010} \\
        & HCCNet-T & 0.295 $\pm$ 0.019 & 0.643 $\pm$ 0.101 & 0.231 $\pm$ 0.013 & 0.165 $\pm$ 0.022 & 0.612 $\pm$ 0.116 & 0.104 $\pm$ 0.012 \\
        T1 DCE-MRI & HCCNet-F & 0.354 $\pm$ 0.039 & 0.595 $\pm$ 0.060 & 0.252 $\pm$ 0.026 & 0.211 $\pm$ 0.051 & 0.732 $\pm$ 0.080 & 0.173 $\pm$ 0.039 \\
        & HCCNet-P & 0.335 $\pm$ 0.044 & \textbf{0.571 $\pm$ 0.059} & 0.240 $\pm$ 0.031 & 0.234 $\pm$ 0.066 & 0.693 $\pm$ 0.075 & 0.184 $\pm$ 0.042 \\
        & HCCNet-N & 0.319 $\pm$ 0.022 & 0.668 $\pm$ 0.133 & 0.230 $\pm$ 0.017 & 0.252 $\pm$ 0.090 & 0.707 $\pm$ 0.120 & 0.211 $\pm$ 0.052 \\
        & HCCNet-T & 0.311 $\pm$ 0.040 & 0.659 $\pm$ 0.113 & 0.220 $\pm$ 0.035 & \textbf{0.112 $\pm$ 0.020} & 0.679 $\pm$ 0.115 & 0.118 $\pm$ 0.012 \\
        T1 IOP \& T2-MRI & HCCNet-F & 0.186 $\pm$ 0.022 & 0.588 $\pm$ 0.096 & 0.168 $\pm$ 0.016 & 0.279 $\pm$ 0.070 & 0.842 $\pm$ 0.079 & 0.276 $\pm$ 0.056 \\
        & HCCNet-P & \textbf{0.158 $\pm$ 0.019} & 0.665 $\pm$ 0.131 & \textbf{0.150 $\pm$ 0.008} & 0.242 $\pm$ 0.047 & 0.791 $\pm$ 0.141 & 0.218 $\pm$ 0.035 \\
        & HCCNet-N & \underline{0.165 $\pm$ 0.022} & 0.653 $\pm$ 0.133 & \underline{0.161 $\pm$ 0.012} & 0.246 $\pm$ 0.033 & 0.647 $\pm$ 0.073 & 0.210 $\pm$ 0.018 \\
        & HCCNet-T & 0.174 $\pm$ 0.032 & \underline{0.582 $\pm$ 0.114} & 0.164 $\pm$ 0.017 & 0.261 $\pm$ 0.049 & 0.644 $\pm$ 0.125 & 0.219 $\pm$ 0.022 \\
        \bottomrule
    \end{tabular}
    \begin{tablenotes}
        \Large
        \item \textit{Note:} Reliability metrics range from 0 to 1. ECE represents the expected, while MCE denotes the maximal calibration error with lower values representing better model performance. All metrics are presented using the mean and standard deviation across the individual 10 model runs. The best model results are emphasised in boldface. Second best results are underlined. DW-MRI comprises diffusion MRI scans with diffusion coefficient $b \in \{0, 150, 400, 800\}$. T1 DCE-MRI contain a pre-contrast and three post-contrast (i.e., late aterial, portal venous, and delayed phase) series. T1 IOP \& T2-MRI comprise T1-weighted in- and out-of-phase MRI as well as T2-weighted MRI with short and long echo times.
    \end{tablenotes}
    \end{threeparttable}
    }
    \normalsize
\end{table}

Table \ref{tab:reliability_results} presents the expected and maximal calibration error as well as the Brier score for baseline and fine-tuned model results stratified by imaging modality and model architecture. As for the comparison across performance metrics depicted in Table \ref{tab:results_base_pre}, variations in reliability scores are predominantly driven by imaging modality rather than model complexity. For the baseline model, models trained using in- and out-of-phase T1-weighted and T2-weighted MRIs consistently display the lowest average reliability scores. That is despite that baseline models trained using dynamic contrast enhanced T1-weighted images show the best average discriminatory power between patients of low and high risk of HCC. Contrarily, for the fine-tuned models, preceding pre-training establishes a negative relationship between performance and reliability metrics. Here, HCCNet fine-tuned using diffusion weighted MRIs display the lowest average reliability scores, while models fine-tuned using T1 \& T2-weighted MR images exhibit the highest average reliability metrics.

\clearpage
\bibliographystyle{layout/elsarticle-harv}

\begin{thebibliography}{32}
\expandafter\ifx\csname natexlab\endcsname\relax\def\natexlab#1{#1}\fi
\providecommand{\url}[1]{\texttt{#1}}
\providecommand{\href}[2]{#2}
\providecommand{\path}[1]{#1}
\providecommand{\DOIprefix}{doi:}
\providecommand{\ArXivprefix}{arXiv:}
\providecommand{\URLprefix}{URL: }
\providecommand{\Pubmedprefix}{pmid:}
\providecommand{\doi}[1]{\href{http://dx.doi.org/#1}{\path{#1}}}
\providecommand{\Pubmed}[1]{\href{pmid:#1}{\path{#1}}}
\providecommand{\bibinfo}[2]{#2}
\ifx\xfnm\relax \def\xfnm[#1]{\unskip,\space#1}\fi
\bibitem[{Caron et~al.(2021)Caron, Touvron, Misra, Jégou, Mairal, Bojanowski and Joulin}]{caronEmergingPropertiesSelfSupervised2021}
\bibinfo{author}{Caron, M.}, \bibinfo{author}{Touvron, H.}, \bibinfo{author}{Misra, I.}, \bibinfo{author}{Jégou, H.}, \bibinfo{author}{Mairal, J.}, \bibinfo{author}{Bojanowski, P.}, \bibinfo{author}{Joulin, A.}, \bibinfo{year}{2021}.
\newblock \bibinfo{title}{Emerging {Properties} in {Self}-{Supervised} {Vision} {Transformers}}.
\newblock \URLprefix \url{http://arxiv.org/abs/2104.14294}. \bibinfo{note}{arXiv:2104.14294 [cs]}.
\bibitem[{Chen et~al.(2020)Chen, Kornblith, Norouzi and Hinton}]{chenSimpleFrameworkContrastive2020}
\bibinfo{author}{Chen, T.}, \bibinfo{author}{Kornblith, S.}, \bibinfo{author}{Norouzi, M.}, \bibinfo{author}{Hinton, G.}, \bibinfo{year}{2020}.
\newblock \bibinfo{title}{A {Simple} {Framework} for {Contrastive} {Learning} of {Visual} {Representations}}.
\newblock \URLprefix \url{http://arxiv.org/abs/2002.05709}. \bibinfo{note}{arXiv:2002.05709 [cs, stat]}.
\bibitem[{Dadsetan et~al.(2022)Dadsetan, Arefan, Berg, Zuley, Sumkin and Wu}]{dadsetanDeepLearningLongitudinal2022}
\bibinfo{author}{Dadsetan, S.}, \bibinfo{author}{Arefan, D.}, \bibinfo{author}{Berg, W.A.}, \bibinfo{author}{Zuley, M.L.}, \bibinfo{author}{Sumkin, J.H.}, \bibinfo{author}{Wu, S.}, \bibinfo{year}{2022}.
\newblock \bibinfo{title}{Deep learning of longitudinal mammogram examinations for breast cancer risk prediction}.
\newblock \bibinfo{journal}{Pattern Recognition} \bibinfo{volume}{132}, \bibinfo{pages}{108919}.
\newblock \URLprefix \url{https://www.sciencedirect.com/science/article/pii/S0031320322004009}, \DOIprefix\doi{10.1016/j.patcog.2022.108919}.
\bibitem[{Dammu et~al.(2023)Dammu, Ren and Duong}]{dammuDeepLearningPrediction2023}
\bibinfo{author}{Dammu, H.}, \bibinfo{author}{Ren, T.}, \bibinfo{author}{Duong, T.Q.}, \bibinfo{year}{2023}.
\newblock \bibinfo{title}{Deep learning prediction of pathological complete response, residual cancer burden, and progression-free survival in breast cancer patients}.
\newblock \bibinfo{journal}{PLOS ONE} \bibinfo{volume}{18}, \bibinfo{pages}{e0280148}.
\newblock \URLprefix \url{https://journals.plos.org/plosone/article?id=10.1371/journal.pone.0280148}, \DOIprefix\doi{10.1371/journal.pone.0280148}. \bibinfo{note}{publisher: Public Library of Science}.
\bibitem[{El-Nouby et~al.(2021)El-Nouby, Izacard, Touvron, Laptev, Jegou and Grave}]{el-noubyAreLargescaleDatasets2021}
\bibinfo{author}{El-Nouby, A.}, \bibinfo{author}{Izacard, G.}, \bibinfo{author}{Touvron, H.}, \bibinfo{author}{Laptev, I.}, \bibinfo{author}{Jegou, H.}, \bibinfo{author}{Grave, E.}, \bibinfo{year}{2021}.
\newblock \bibinfo{title}{Are {Large}-scale {Datasets} {Necessary} for {Self}-{Supervised} {Pre}-training?}
\newblock \URLprefix \url{http://arxiv.org/abs/2112.10740}, \DOIprefix\doi{10.48550/arXiv.2112.10740}. \bibinfo{note}{arXiv:2112.10740 [cs]}.
\bibitem[{Gao et~al.(2021)Gao, Ghodrati, Kalbasi, Fu, Ruan, Cao, Wang, Eilber, Bernthal, Bukata, Dry, Nelson, Kamrava, Lewis, Low, Steinberg, Hu and Yang}]{gaoPredictionSoftTissue2021a}
\bibinfo{author}{Gao, Y.}, \bibinfo{author}{Ghodrati, V.}, \bibinfo{author}{Kalbasi, A.}, \bibinfo{author}{Fu, J.}, \bibinfo{author}{Ruan, D.}, \bibinfo{author}{Cao, M.}, \bibinfo{author}{Wang, C.}, \bibinfo{author}{Eilber, F.C.}, \bibinfo{author}{Bernthal, N.}, \bibinfo{author}{Bukata, S.}, \bibinfo{author}{Dry, S.M.}, \bibinfo{author}{Nelson, S.D.}, \bibinfo{author}{Kamrava, M.}, \bibinfo{author}{Lewis, J.}, \bibinfo{author}{Low, D.A.}, \bibinfo{author}{Steinberg, M.}, \bibinfo{author}{Hu, P.}, \bibinfo{author}{Yang, Y.}, \bibinfo{year}{2021}.
\newblock \bibinfo{title}{Prediction of soft tissue sarcoma response to radiotherapy using longitudinal diffusion {MRI} and a deep neural network with generative adversarial network-based data augmentation}.
\newblock \bibinfo{journal}{Medical Physics} \bibinfo{volume}{48}, \bibinfo{pages}{3262--3372}.
\newblock \URLprefix \url{https://onlinelibrary.wiley.com/doi/abs/10.1002/mp.14897}, \DOIprefix\doi{10.1002/mp.14897}. \bibinfo{note}{\_eprint: https://onlinelibrary.wiley.com/doi/pdf/10.1002/mp.14897}.
\bibitem[{Gong et~al.(2024)Gong, Beckmann and Smith}]{gong2024individualised}
\bibinfo{author}{Gong, W.}, \bibinfo{author}{Beckmann, C.F.}, \bibinfo{author}{Smith, S.M.}, \bibinfo{year}{2024}.
\newblock \bibinfo{title}{Individualised prediction of longitudinal change in multimodal brain imaging}.
\newblock \bibinfo{journal}{Imaging Neuroscience} \bibinfo{volume}{2}, \bibinfo{pages}{1--19}.
\bibitem[{Granziol et~al.()Granziol, Zohren and Roberts}]{granziolLearningRatesFunction}
\bibinfo{author}{Granziol, D.}, \bibinfo{author}{Zohren, S.}, \bibinfo{author}{Roberts, S.}, .
\newblock \bibinfo{title}{Learning {Rates} as a {Function} of {Batch} {Size}: {A} {Random} {Matrix} {Theory} {Approach} to {Neural} {Network} {Training}} .
\bibitem[{Grill et~al.(2020)Grill, Strub, Altché, Tallec, Richemond, Buchatskaya, Doersch, Pires, Guo, Azar, Piot, Kavukcuoglu, Munos and Valko}]{grillBootstrapYourOwn2020}
\bibinfo{author}{Grill, J.B.}, \bibinfo{author}{Strub, F.}, \bibinfo{author}{Altché, F.}, \bibinfo{author}{Tallec, C.}, \bibinfo{author}{Richemond, P.H.}, \bibinfo{author}{Buchatskaya, E.}, \bibinfo{author}{Doersch, C.}, \bibinfo{author}{Pires, B.A.}, \bibinfo{author}{Guo, Z.D.}, \bibinfo{author}{Azar, M.G.}, \bibinfo{author}{Piot, B.}, \bibinfo{author}{Kavukcuoglu, K.}, \bibinfo{author}{Munos, R.}, \bibinfo{author}{Valko, M.}, \bibinfo{year}{2020}.
\newblock \bibinfo{title}{Bootstrap your own latent: {A} new approach to self-supervised {Learning}}.
\newblock \URLprefix \url{http://arxiv.org/abs/2006.07733}. \bibinfo{note}{arXiv:2006.07733 [cs, stat]}.
\bibitem[{Huang et~al.(2023)Huang, Pareek, Jensen, Lungren, Yeung and Chaudhari}]{huangSelfsupervisedLearningMedical2023}
\bibinfo{author}{Huang, S.C.}, \bibinfo{author}{Pareek, A.}, \bibinfo{author}{Jensen, M.}, \bibinfo{author}{Lungren, M.P.}, \bibinfo{author}{Yeung, S.}, \bibinfo{author}{Chaudhari, A.S.}, \bibinfo{year}{2023}.
\newblock \bibinfo{title}{Self-supervised learning for medical image classification: a systematic review and implementation guidelines}.
\newblock \bibinfo{journal}{npj Digital Medicine} \bibinfo{volume}{6}, \bibinfo{pages}{1--16}.
\newblock \URLprefix \url{https://www.nature.com/articles/s41746-023-00811-0}, \DOIprefix\doi{10.1038/s41746-023-00811-0}. \bibinfo{note}{publisher: Nature Publishing Group}.
\bibitem[{Jamaludin et~al.(2017)Jamaludin, Kadir and Zisserman}]{jamaludinSelfsupervisedLearningSpinal2017}
\bibinfo{author}{Jamaludin, A.}, \bibinfo{author}{Kadir, T.}, \bibinfo{author}{Zisserman, A.}, \bibinfo{year}{2017}.
\newblock \bibinfo{title}{Self-supervised {Learning} for {Spinal} {MRIs}}, in: \bibinfo{editor}{Cardoso, M.J.}, \bibinfo{editor}{Arbel, T.}, \bibinfo{editor}{Carneiro, G.}, \bibinfo{editor}{Syeda-Mahmood, T.}, \bibinfo{editor}{Tavares, J.M.R.}, \bibinfo{editor}{Moradi, M.}, \bibinfo{editor}{Bradley, A.}, \bibinfo{editor}{Greenspan, H.}, \bibinfo{editor}{Papa, J.P.}, \bibinfo{editor}{Madabhushi, A.}, \bibinfo{editor}{Nascimento, J.C.}, \bibinfo{editor}{Cardoso, J.S.}, \bibinfo{editor}{Belagiannis, V.}, \bibinfo{editor}{Lu, Z.} (Eds.), \bibinfo{booktitle}{Deep {Learning} in {Medical} {Image} {Analysis} and {Multimodal} {Learning} for {Clinical} {Decision} {Support}}, \bibinfo{publisher}{Springer International Publishing}, \bibinfo{address}{Cham}. pp. \bibinfo{pages}{294--302}.
\newblock \DOIprefix\doi{10.1007/978-3-319-67558-9_34}.
\bibitem[{Jin et~al.(2021a)Jin, Yu, Ke, Ding, Yi, Jiang, Duan, Tang, Chang, Wu, Gao and Li}]{jinPredictingTreatmentResponse2021}
\bibinfo{author}{Jin, C.}, \bibinfo{author}{Yu, H.}, \bibinfo{author}{Ke, J.}, \bibinfo{author}{Ding, P.}, \bibinfo{author}{Yi, Y.}, \bibinfo{author}{Jiang, X.}, \bibinfo{author}{Duan, X.}, \bibinfo{author}{Tang, J.}, \bibinfo{author}{Chang, D.T.}, \bibinfo{author}{Wu, X.}, \bibinfo{author}{Gao, F.}, \bibinfo{author}{Li, R.}, \bibinfo{year}{2021}a.
\newblock \bibinfo{title}{Predicting treatment response from longitudinal images using multi-task deep learning}.
\newblock \bibinfo{journal}{Nature Communications} \bibinfo{volume}{12}, \bibinfo{pages}{1851}.
\newblock \URLprefix \url{https://www.nature.com/articles/s41467-021-22188-y}, \DOIprefix\doi{10.1038/s41467-021-22188-y}. \bibinfo{note}{publisher: Nature Publishing Group}.
\bibitem[{Jin et~al.(2021b)Jin, Yu, Ke, Ding, Yi, Jiang, Duan, Tang, Chang, Wu et~al.}]{jin2021predicting}
\bibinfo{author}{Jin, C.}, \bibinfo{author}{Yu, H.}, \bibinfo{author}{Ke, J.}, \bibinfo{author}{Ding, P.}, \bibinfo{author}{Yi, Y.}, \bibinfo{author}{Jiang, X.}, \bibinfo{author}{Duan, X.}, \bibinfo{author}{Tang, J.}, \bibinfo{author}{Chang, D.T.}, \bibinfo{author}{Wu, X.}, et~al., \bibinfo{year}{2021}b.
\newblock \bibinfo{title}{Predicting treatment response from longitudinal images using multi-task deep learning}.
\newblock \bibinfo{journal}{Nature Communications} \bibinfo{volume}{12}, \bibinfo{pages}{1--12}.
\bibitem[{Kumar and Marttinen(2024)}]{kumar2024improving}
\bibinfo{author}{Kumar, Y.}, \bibinfo{author}{Marttinen, P.}, \bibinfo{year}{2024}.
\newblock \bibinfo{title}{Improving medical multi-modal contrastive learning with expert annotations}.
\newblock \bibinfo{journal}{arXiv preprint arXiv:2403.10153} .
\bibitem[{Lan et~al.(2020)Lan, Chen, Goodman, Gimpel, Sharma and Soricut}]{lanALBERTLiteBERT2020}
\bibinfo{author}{Lan, Z.}, \bibinfo{author}{Chen, M.}, \bibinfo{author}{Goodman, S.}, \bibinfo{author}{Gimpel, K.}, \bibinfo{author}{Sharma, P.}, \bibinfo{author}{Soricut, R.}, \bibinfo{year}{2020}.
\newblock \bibinfo{title}{{ALBERT}: {A} {Lite} {BERT} for {Self}-supervised {Learning} of {Language} {Representations}}.
\newblock \URLprefix \url{http://arxiv.org/abs/1909.11942}. \bibinfo{note}{arXiv:1909.11942 [cs]}.
\bibitem[{Lee et~al.(2022)Lee, Hu, Kuo, Alam, Yorke, Li, Rimner and Zhang}]{leeDeepLearningDriven2022}
\bibinfo{author}{Lee, D.}, \bibinfo{author}{Hu, Y.c.}, \bibinfo{author}{Kuo, L.}, \bibinfo{author}{Alam, S.}, \bibinfo{author}{Yorke, E.}, \bibinfo{author}{Li, A.}, \bibinfo{author}{Rimner, A.}, \bibinfo{author}{Zhang, P.}, \bibinfo{year}{2022}.
\newblock \bibinfo{title}{Deep learning driven predictive treatment planning for adaptive radiotherapy of lung cancer}.
\newblock \bibinfo{journal}{Radiotherapy and Oncology} \bibinfo{volume}{169}, \bibinfo{pages}{57--63}.
\newblock \URLprefix \url{https://www.thegreenjournal.com/article/S0167-8140(22)00098-6/fulltext}, \DOIprefix\doi{10.1016/j.radonc.2022.02.013}. \bibinfo{note}{publisher: Elsevier}.
\bibitem[{Li et~al.(2020)Li, Chen and Yang}]{liUnderstandingDisharmonyWeight2020}
\bibinfo{author}{Li, X.}, \bibinfo{author}{Chen, S.}, \bibinfo{author}{Yang, J.}, \bibinfo{year}{2020}.
\newblock \bibinfo{title}{Understanding the disharmony between weight normalization family and weight decay}, in: \bibinfo{booktitle}{Proceedings of the {AAAI} {Conference} on {Artificial} {Intelligence}}, pp. \bibinfo{pages}{4715--4722}.
\newblock \URLprefix \url{https://ojs.aaai.org/index.php/AAAI/article/view/5904}. \bibinfo{note}{issue: 04}.
\bibitem[{Liu et~al.(2022)Liu, Mao, Wu, Feichtenhofer, Darrell and Xie}]{liuConvNet2020s2022}
\bibinfo{author}{Liu, Z.}, \bibinfo{author}{Mao, H.}, \bibinfo{author}{Wu, C.Y.}, \bibinfo{author}{Feichtenhofer, C.}, \bibinfo{author}{Darrell, T.}, \bibinfo{author}{Xie, S.}, \bibinfo{year}{2022}.
\newblock \bibinfo{title}{A {ConvNet} for the 2020s}.
\newblock \URLprefix \url{http://arxiv.org/abs/2201.03545}. \bibinfo{note}{arXiv:2201.03545 [cs]}.
\bibitem[{Loshchilov and Hutter(2019)}]{loshchilovDecoupledWeightDecay2019}
\bibinfo{author}{Loshchilov, I.}, \bibinfo{author}{Hutter, F.}, \bibinfo{year}{2019}.
\newblock \bibinfo{title}{Decoupled {Weight} {Decay} {Regularization}}.
\newblock \URLprefix \url{http://arxiv.org/abs/1711.05101}, \DOIprefix\doi{10.48550/arXiv.1711.05101}. \bibinfo{note}{arXiv:1711.05101 [cs, math]}.
\bibitem[{Lu et~al.(2021)Lu, Dercle, Zhao and Schwartz}]{luDeepLearningPrediction2021}
\bibinfo{author}{Lu, L.}, \bibinfo{author}{Dercle, L.}, \bibinfo{author}{Zhao, B.}, \bibinfo{author}{Schwartz, L.H.}, \bibinfo{year}{2021}.
\newblock \bibinfo{title}{Deep learning for the prediction of early on-treatment response in metastatic colorectal cancer from serial medical imaging}.
\newblock \bibinfo{journal}{Nature Communications} \bibinfo{volume}{12}, \bibinfo{pages}{6654}.
\newblock \URLprefix \url{https://www.nature.com/articles/s41467-021-26990-6}, \DOIprefix\doi{10.1038/s41467-021-26990-6}. \bibinfo{note}{publisher: Nature Publishing Group}.
\bibitem[{Parikh et~al.(2020)Parikh, Mehta, Singal, Block, Marrero and Lok}]{parikh2020biomarkers}
\bibinfo{author}{Parikh, N.D.}, \bibinfo{author}{Mehta, A.S.}, \bibinfo{author}{Singal, A.G.}, \bibinfo{author}{Block, T.}, \bibinfo{author}{Marrero, J.A.}, \bibinfo{author}{Lok, A.S.}, \bibinfo{year}{2020}.
\newblock \bibinfo{title}{Biomarkers for the early detection of hepatocellular carcinoma}.
\newblock \bibinfo{journal}{Cancer Epidemiology, Biomarkers \& Prevention} \bibinfo{volume}{29}, \bibinfo{pages}{2495--2503}.
\bibitem[{Ren et~al.(2021)Ren, Wang, Zhao and Wu}]{renRAPTPretrainingTimeAware2021}
\bibinfo{author}{Ren, H.}, \bibinfo{author}{Wang, J.}, \bibinfo{author}{Zhao, W.X.}, \bibinfo{author}{Wu, N.}, \bibinfo{year}{2021}.
\newblock \bibinfo{title}{{RAPT}: {Pre}-training of {Time}-{Aware} {Transformer} for {Learning} {Robust} {Healthcare} {Representation}}, in: \bibinfo{booktitle}{Proceedings of the 27th {ACM} {SIGKDD} {Conference} on {Knowledge} {Discovery} \& {Data} {Mining}}, \bibinfo{publisher}{ACM}, \bibinfo{address}{Virtual Event Singapore}. pp. \bibinfo{pages}{3503--3511}.
\newblock \URLprefix \url{https://dl.acm.org/doi/10.1145/3447548.3467069}, \DOIprefix\doi{10.1145/3447548.3467069}.
\bibitem[{Rivail et~al.(2019)Rivail, Schmidt-Erfurth, Vogl, Waldstein, Riedl, Grechenig, Wu and Bogunovic}]{rivailModelingDiseaseProgression2019}
\bibinfo{author}{Rivail, A.}, \bibinfo{author}{Schmidt-Erfurth, U.}, \bibinfo{author}{Vogl, W.D.}, \bibinfo{author}{Waldstein, S.M.}, \bibinfo{author}{Riedl, S.}, \bibinfo{author}{Grechenig, C.}, \bibinfo{author}{Wu, Z.}, \bibinfo{author}{Bogunovic, H.}, \bibinfo{year}{2019}.
\newblock \bibinfo{title}{Modeling {Disease} {Progression} in {Retinal} {OCTs} with {Longitudinal} {Self}-supervised {Learning}}, in: \bibinfo{editor}{Rekik, I.}, \bibinfo{editor}{Adeli, E.}, \bibinfo{editor}{Park, S.H.} (Eds.), \bibinfo{booktitle}{Predictive {Intelligence} in {Medicine}}, \bibinfo{publisher}{Springer International Publishing}, \bibinfo{address}{Cham}. pp. \bibinfo{pages}{44--52}.
\newblock \DOIprefix\doi{10.1007/978-3-030-32281-6_5}.
\bibitem[{Tang et~al.(2021)Tang, Yang, Li, Roth, Landman, Xu, Nath and Hatamizadeh}]{tang2021self}
\bibinfo{author}{Tang, Y.}, \bibinfo{author}{Yang, D.}, \bibinfo{author}{Li, W.}, \bibinfo{author}{Roth, H.}, \bibinfo{author}{Landman, B.}, \bibinfo{author}{Xu, D.}, \bibinfo{author}{Nath, V.}, \bibinfo{author}{Hatamizadeh, A.}, \bibinfo{year}{2021}.
\newblock \bibinfo{title}{Self-supervised pre-training of swin transformers for 3d medical image analysis}.
\newblock \bibinfo{journal}{arXiv preprint arXiv:2111.14791} .
\bibitem[{Vaswani et~al.(2023)Vaswani, Shazeer, Parmar, Uszkoreit, Jones, Gomez, Kaiser and Polosukhin}]{vaswaniAttentionAllYou2023}
\bibinfo{author}{Vaswani, A.}, \bibinfo{author}{Shazeer, N.}, \bibinfo{author}{Parmar, N.}, \bibinfo{author}{Uszkoreit, J.}, \bibinfo{author}{Jones, L.}, \bibinfo{author}{Gomez, A.N.}, \bibinfo{author}{Kaiser, L.}, \bibinfo{author}{Polosukhin, I.}, \bibinfo{year}{2023}.
\newblock \bibinfo{title}{Attention {Is} {All} {You} {Need}}.
\newblock \URLprefix \url{http://arxiv.org/abs/1706.03762}. \bibinfo{note}{arXiv:1706.03762 [cs]}.
\bibitem[{Wang et~al.(2020)Wang, Alam, Zhang, Hu, Nadeem, Tyagi, Rimner, Lu, Thor and Zhang}]{wangPredictingSpatialEsophageal2020}
\bibinfo{author}{Wang, C.}, \bibinfo{author}{Alam, S.R.}, \bibinfo{author}{Zhang, S.}, \bibinfo{author}{Hu, Y.C.}, \bibinfo{author}{Nadeem, S.}, \bibinfo{author}{Tyagi, N.}, \bibinfo{author}{Rimner, A.}, \bibinfo{author}{Lu, W.}, \bibinfo{author}{Thor, M.}, \bibinfo{author}{Zhang, P.}, \bibinfo{year}{2020}.
\newblock \bibinfo{title}{Predicting spatial esophageal changes in a multimodal longitudinal imaging study via a convolutional recurrent neural network}.
\newblock \bibinfo{journal}{Physics in Medicine \& Biology} \bibinfo{volume}{65}, \bibinfo{pages}{235027}.
\newblock \URLprefix \url{https://dx.doi.org/10.1088/1361-6560/abb1d9}, \DOIprefix\doi{10.1088/1361-6560/abb1d9}. \bibinfo{note}{publisher: IOP Publishing}.
\bibitem[{Wang et~al.(2019)Wang, Rimner, Hu, Tyagi, Jiang, Yorke, Riyahi, Mageras, Deasy and Zhang}]{wangPredictingEvolutionLung2019}
\bibinfo{author}{Wang, C.}, \bibinfo{author}{Rimner, A.}, \bibinfo{author}{Hu, Y.C.}, \bibinfo{author}{Tyagi, N.}, \bibinfo{author}{Jiang, J.}, \bibinfo{author}{Yorke, E.}, \bibinfo{author}{Riyahi, S.}, \bibinfo{author}{Mageras, G.}, \bibinfo{author}{Deasy, J.O.}, \bibinfo{author}{Zhang, P.}, \bibinfo{year}{2019}.
\newblock \bibinfo{title}{Toward predicting the evolution of lung tumors during radiotherapy observed on a longitudinal {MR} imaging study via a deep learning algorithm}.
\newblock \bibinfo{journal}{Medical Physics} \bibinfo{volume}{46}, \bibinfo{pages}{4699--4707}.
\newblock \URLprefix \url{https://onlinelibrary.wiley.com/doi/abs/10.1002/mp.13765}, \DOIprefix\doi{10.1002/mp.13765}. \bibinfo{note}{\_eprint: https://onlinelibrary.wiley.com/doi/pdf/10.1002/mp.13765}.
\bibitem[{Xiong et~al.(2020)Xiong, Yang, He, Zheng, Zheng, Xing, Zhang, Lan, Wang and Liu}]{xiongLayerNormalizationTransformer2020}
\bibinfo{author}{Xiong, R.}, \bibinfo{author}{Yang, Y.}, \bibinfo{author}{He, D.}, \bibinfo{author}{Zheng, K.}, \bibinfo{author}{Zheng, S.}, \bibinfo{author}{Xing, C.}, \bibinfo{author}{Zhang, H.}, \bibinfo{author}{Lan, Y.}, \bibinfo{author}{Wang, L.}, \bibinfo{author}{Liu, T.Y.}, \bibinfo{year}{2020}.
\newblock \bibinfo{title}{On {Layer} {Normalization} in the {Transformer} {Architecture}}.
\newblock \URLprefix \url{http://arxiv.org/abs/2002.04745}. \bibinfo{note}{arXiv:2002.04745 [cs, stat]}.
\bibitem[{Xu et~al.(2019)Xu, Hosny, Zeleznik, Parmar, Coroller, Franco, Mak and Aerts}]{xuDeepLearningPredicts2019}
\bibinfo{author}{Xu, Y.}, \bibinfo{author}{Hosny, A.}, \bibinfo{author}{Zeleznik, R.}, \bibinfo{author}{Parmar, C.}, \bibinfo{author}{Coroller, T.}, \bibinfo{author}{Franco, I.}, \bibinfo{author}{Mak, R.H.}, \bibinfo{author}{Aerts, H.J.}, \bibinfo{year}{2019}.
\newblock \bibinfo{title}{Deep {Learning} {Predicts} {Lung} {Cancer} {Treatment} {Response} from {Serial} {Medical} {Imaging}}.
\newblock \bibinfo{journal}{Clinical Cancer Research} \bibinfo{volume}{25}, \bibinfo{pages}{3266--3275}.
\newblock \URLprefix \url{https://doi.org/10.1158/1078-0432.CCR-18-2495}, \DOIprefix\doi{10.1158/1078-0432.CCR-18-2495}.
\bibitem[{Zeng et~al.(2021)Zeng, Wu, Hu, Xu, Yuan, Huang, Zhuang, Hu and Shi}]{zeng2021positional}
\bibinfo{author}{Zeng, D.}, \bibinfo{author}{Wu, Y.}, \bibinfo{author}{Hu, X.}, \bibinfo{author}{Xu, X.}, \bibinfo{author}{Yuan, H.}, \bibinfo{author}{Huang, M.}, \bibinfo{author}{Zhuang, J.}, \bibinfo{author}{Hu, J.}, \bibinfo{author}{Shi, Y.}, \bibinfo{year}{2021}.
\newblock \bibinfo{title}{Positional contrastive learning for volumetric medical image segmentation}.
\newblock \bibinfo{journal}{arXiv preprint arXiv:2106.09157} .
\bibitem[{Zhang et~al.(2022a)Zhang, Li, Zhou, Ma and Yu}]{zhangAdvancing3DMedical2022}
\bibinfo{author}{Zhang, S.}, \bibinfo{author}{Li, Z.}, \bibinfo{author}{Zhou, H.Y.}, \bibinfo{author}{Ma, J.}, \bibinfo{author}{Yu, Y.}, \bibinfo{year}{2022}a.
\newblock \bibinfo{title}{Advancing {3D} {Medical} {Image} {Analysis} with {Variable} {Dimension} {Transform} based {Supervised} {3D} {Pre}-training}.
\newblock \URLprefix \url{http://arxiv.org/abs/2201.01426}. \bibinfo{note}{arXiv:2201.01426 [cs, eess]}.
\bibitem[{Zhang et~al.(2022b)Zhang, Hu, Zhong, Song, Sun, Li, Dai, Zhou and Yang}]{zhangSpatiotemporalAttentionEarly2022}
\bibinfo{author}{Zhang, Y.}, \bibinfo{author}{Hu, C.}, \bibinfo{author}{Zhong, L.}, \bibinfo{author}{Song, Y.}, \bibinfo{author}{Sun, J.}, \bibinfo{author}{Li, M.}, \bibinfo{author}{Dai, L.}, \bibinfo{author}{Zhou, Y.}, \bibinfo{author}{Yang, W.}, \bibinfo{year}{2022}b.
\newblock \bibinfo{title}{Spatiotemporal {Attention} for {Early} {Prediction} of {Hepatocellular} {Carcinoma} {Based} on {Longitudinal} {Ultrasound} {Images}}, in: \bibinfo{editor}{Wang, L.}, \bibinfo{editor}{Dou, Q.}, \bibinfo{editor}{Fletcher, P.T.}, \bibinfo{editor}{Speidel, S.}, \bibinfo{editor}{Li, S.} (Eds.), \bibinfo{booktitle}{Medical {Image} {Computing} and {Computer} {Assisted} {Intervention} – {MICCAI} 2022}, \bibinfo{publisher}{Springer Nature Switzerland}, \bibinfo{address}{Cham}. pp. \bibinfo{pages}{534--543}.
\newblock \DOIprefix\doi{10.1007/978-3-031-16437-8_51}.

\end{thebibliography}

\end{document}